\documentclass{article} 
\usepackage[table,xcdraw]{xcolor}
\usepackage{iclr2022_conference,times}

\iclrfinalcopy

\usepackage{amsmath,amsfonts,bm}

\def\eqref#1{equation~\ref{#1}}

\def\1{\bm{1}}

\DeclareMathAlphabet{\mathsfit}{\encodingdefault}{\sfdefault}{m}{sl}
\SetMathAlphabet{\mathsfit}{bold}{\encodingdefault}{\sfdefault}{bx}{n}

\newcommand{\R}{\mathbb{R}}

\usepackage{latexsym}

\usepackage{url}
\usepackage[T1]{fontenc}

\usepackage{graphicx}
\usepackage{verbatim}
\usepackage{booktabs}       
\usepackage{amsfonts}       
\usepackage{nicefrac}       
\usepackage{amsmath,amsfonts,bm}
\usepackage{hyperref}
\usepackage{subcaption}
\usepackage{pdflscape}

\usepackage{float}
\usepackage{longtable}

\usepackage{tikz}
\usetikzlibrary{external,positioning,fit,shapes,arrows,arrows.meta,patterns}
\usepackage{pgfplots}
\usepackage{pgfplotstable}
\usepgfplotslibrary{colorbrewer}
\pgfplotsset{compat=newest}

\usepackage{catchfile}
\usepackage{ifthen}
\usepgfplotslibrary{fillbetween}

\hypersetup{
    colorlinks=false,
    linkcolor=blue,
    filecolor=magenta,      
    urlcolor=cyan,
}

\def\acc{\text{acc}}

\def\x{\mathbf{x}}
\def\y{\mathbf{y}}
\def\rnn{\text{RNN}}
\def\entropy{H}
\def\R{\mathbb{R}}
\def\vec{\text{vec}}
\def\sys/{\textsc{Icy}}

\newcommand\teal[1]{\textcolor{teal}{#1}}
\newcommand\purple[1]{\textcolor{purple}{#1}}
\newcommand\orange[1]{\textcolor{orange}{#1}}
\newcommand\blue[1]{\textcolor{blue}{#1}}

\pgfplotsset{
    discard if not/.style 2 args={
        x filter/.code={
            \edef\tempa{\thisrow{#1}}
            \edef\tempb{#2}
            \ifx\tempa\tempb
            \else
                
            \fi
        }
    }
}

\newenvironment{customlegend}[1][]{%
    \begingroup
    \csname pgfplots@init@cleared@structures\endcsname
    \pgfplotsset{#1}%
}{%
    \csname pgfplots@createlegend\endcsname
    \endgroup
}%

\def\addlegendimage{\csname pgfplots@addlegendimage\endcsname}

\pgfplotsset{
    discard if not two/.style n args={4}{
        x filter/.code={
            \edef\tempa{\thisrow{#1}}
            \edef\tempb{#2}
            \ifx\tempa\tempb
                \edef\tempa{\thisrow{#3}}
                \edef\tempb{#4}
                \ifx\tempa\tempb
                \else
                    
                \fi
            \else
                
            \fi
        }
    }
}

\tikzstyle{myarrow} = [>={Stealth[round]},shorten >=1pt,semithick]
\tikzstyle{block} = [draw, inner sep=0.2cm, node distance=0.6cm, fill=yellow!10, font=\small]
\tikzstyle{m} = [draw, inner sep=0.2cm, node distance=0.6cm, fill=blue!10, font=\small]
\tikzstyle{u} = [draw, inner sep=0.2cm, node distance=0.6cm, fill=green!10, font=\small]
\tikzstyle{xblock} = [draw, inner sep=0.2cm, node distance=0.6cm, fill=red!10, font=\small]
\tikzstyle{gray} = [draw, inner sep=0.2cm, node distance=0.6cm, fill=black!03, font=\small]

\title{Neural networks can understand compositional functions that humans do not, in the context of emergent communication}

\author{{ Hugh Perkins (hp@asapp.com)} \\
  ASAPP (https://asapp.com) \\
  1 World Trade Center, NY 10007 USA
 }

\date{}

\begin{document}

\maketitle

\begin{abstract}
We show that it is possible to craft transformations that, applied to compositional grammars, result in grammars that neural networks can learn easily, but humans do not.
This could explain the disconnect between current metrics of compositionality, that are arguably human-centric, and the ability of neural networks to generalize to unseen examples.
We propose to use the transformations as a benchmark, \textsc{Icy}, which could be used to measure aspects of the compositional inductive bias of networks, and to search for networks with similar compositional inductive biases to humans.
As an example of this approach, we propose a hierarchical model, HU-RNN, which shows an inductive bias towards position-independent, word-like groups of tokens.
\end{abstract}


\section{Introduction} \label{section:introduction}

Statistical association language models produce impressive results in domains such as summarization, and few-shot learning (e.g. \citet{zhang2020pegasus}, or \citet{brown2020language}).
However, it is unclear to what extent such tasks require creative invention by the neural models.
Thus, we target a slightly different task of `emergent communication'.
Tabula rasa agents placed in a collaborative scenario emerge their own communicative code (e.g. \citet{lazaridou2018_refgames} and \citet{foerster_multi_agent_rl_2016}).
We wish to reproduce aspects of the development of human natural language (e.g. \citet{pinker1990natural}, \citet{berwick2012bird}).
A key aspect is compositionality: the meaning of an utterance is a function of the meaning of the parts.
Agents in emergent communication scenarios empirically do not naturally produce compositional output, as measured by human evaluation, and by compositional metrics, e.g. \citet{kottur2017natural}.

\citet{kirby2008cumulative} showed in human experiments that artificial languages evolved to become more compositional when transmitted from one human to another.
However, in the case of artificial models, \citet{griffiths2007language} showed that for a broad range of conditions, transmission of languages across generations converges to the prior.
For artificial models, a key question thus is: what are the priors? To what extent do commonly used models incorporate a compositional inductive bias?

\begin{figure}
    \centering
    \includegraphics[width=0.7\textwidth]{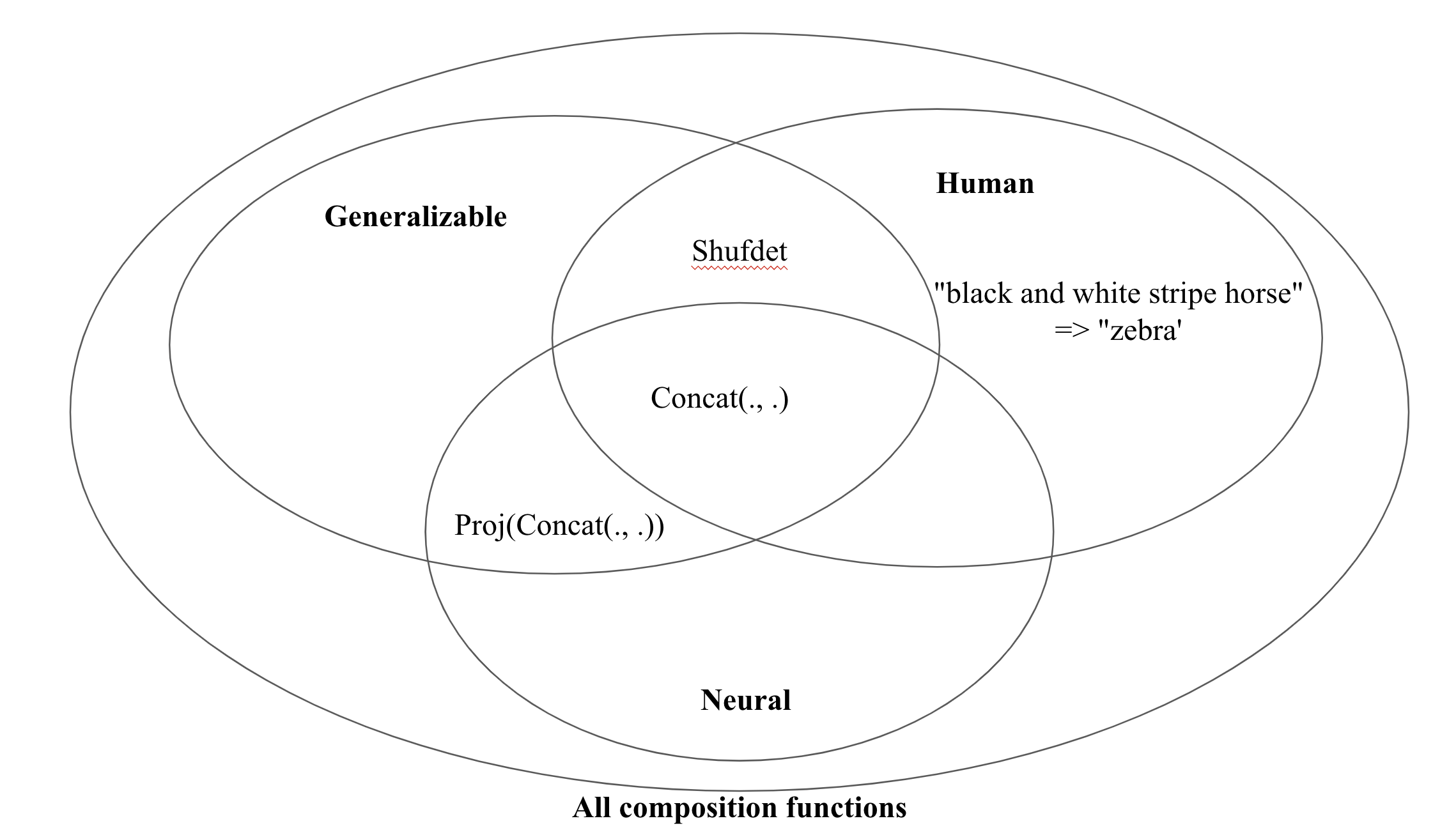}
    \caption{Subsets of composition space}
    \label{fig:composition_venn}
\end{figure}

To go further, we need a concrete definition of compositionality. We use the definition of compositionality from \citet{andreas_measuring_compositionality_2019}:
an utterance representing the combined meaning of two sub-utterances should be a deterministic function $g(\cdot, \cdot)$ of the two sub-utterances. 
This is a broad definition of compositionality, and includes holistic mappings, which do not generalize.
We thus consider two subsets of compositionality, which we term `generalizable' compositionality, and `human' compositionality. Human compositionality is defined to be compositional functions which can be used by humans. Generalizable composition is defined to be any composition function which allows generalization.
Figure \ref{fig:composition_venn} depicts these subsets of composition space, as well as a subset `neural', depicting composition functions usable by current neural models.

Our current metrics of composition implicitly target human compositionality. We hypothesize that a consistently observed disconnect between the measured compositionality of emergent communication grammars, and their ability to generalize \citep{chaabouni2020compositionality}, is a direct consequence of our metrics of compositionality targeting human compositionality.

We present specific examples of generalizable composition functions, which neural models can acquire easily, but which humans do not recognize as compositional, and which current compositional metrics consider to be non-compositional. In addition, we present a grammar, \textsc{shufdet}, whose composition humans can understand but which neural models cannot. We propose a novel neural architecture, HU-RNN, that can acquire \textsc{shufdet} faster than other neural models.

What we can learn from this is three-fold. Firstly, when we talk about compositionality, we should be clear about whether we mean human compositionality, generalizable compositionality, or some other kind of compositionality. Secondly, we should be clear about what our goal is when we wish for emergent communication games to emerge compositional language. Is our goal to make the language appear compositional to humans, or simply that the language appear compositional to neural networks?
Thirdly the compositional inductive bias of current neural networks is quite different from that of humans. There are generalizable compositions that neural networks can use that humans cannot; and 
similarly there are compositional functions, e.g \textsc{shufdet}, that humans can use that current neural networks do not.

Our contributions are:
\begin{itemize}
    \item demonstrate transformations, which we can apply to concatenation grammars which give rise to grammars whose compositional structure:
    \begin{itemize}
        \item appears to current metrics of compositionality as non-compositional
        \item is opaque to humans
        \item does not affect acquisition speed of neural models
    \end{itemize}
    \item we measure the performance of these transformations:
    \begin{itemize}
        \item using current compositional metrics
        \item using human evaluation
        \item using a selection of standard neural models
    \end{itemize}
    \item in addition we propose a transformation, \textsc{shufdet}, which we show that humans can readily understand, but which neural models acquire slowly
    \item as an example of using our transformations to search for models with a compositional inductive bias more aligned with that of humans, we propose a model, HU-RNN, that shows faster acquisition speed for \textsc{shufdet}
\end{itemize}





\section{Background}

\subsection{General Framework}

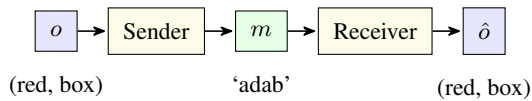
\begin{figure}
\centering
\begin{tikzpicture}[auto]
    \draw[draw=none, use as bounding box] (-2.1, -1.0) rectangle (5.3, 0.4);

    \node [block]                    (sender)   {Sender};
    \node [m, node distance=0.4cm, left=of sender]        (m)        {$o$};
    \node [u, node distance=0.4cm, right=of sender]       (u_pred)   {$m$};
    \node [block, node distance=0.4cm, right=of u_pred]   (receiver) {Receiver};
    \node [m, node distance=0.4cm, right=of receiver]     (m_pred)   {$\hat{o}$};

    \node [node distance=0.2cm, below=of u_pred, font=\small] {`adab'};
    \node [node distance=0.2cm, below=of m, font=\small] {(red, box)};
    \node [node distance=0.2cm, below=of m_pred, font=\small] {(red, box)};


    \draw [myarrow,->] (m) -- (sender);
    \draw [myarrow,->] (sender) -- (u_pred);
    \draw [myarrow,->] (u_pred) -- (receiver);
    \draw [myarrow,->] (receiver) -- (m_pred);

\end{tikzpicture}
\vspace{-0.2cm}
\caption{Signaling Game. `adab' is an example message. (red, box) is an example object.}
\label{fig:signaling_game}
\end{figure}

We assume a signaling game \citep{lewis2008convention}. A Sender receives an object $o$, and generates a message $m$, Figure \ref{fig:signaling_game}. A Receiver receives the message $m$, and decodes the message into a prediction $\hat{o}$ of the original object $o$. The message $m$ is a fixed length utterance of $c_{len}$ symbols drawn from a vocabulary $V$ of size $|V|$. Each object $o$ comprises $n_{att}$ attributes, $\{o^{(1)}, \dots, o^{(n_{att})}\}$ each with $n_{val}$ possible values. We draw the attributes $o^{(j)}$ from a vocabulary $\Sigma$, of size $|\Sigma| = n_{att} \cdot n_{val}$, where
$\Sigma_{j,*}$ are possible values for attribute $o^{(j)}$.  For example, $\Sigma_{1,*}$ could represent color; $|\Sigma_{1,*}| = n_{val}$ would be the number of possible colors; and $\Sigma_{1,3}$ could mean `red'. When presented to a neural network, $o$ is represented as the concatenation of $n_{att}$ one-hot vectors, each of length $n_{val}$.

In emergent communication games, we can ssign a reward $r=1$ if $\hat{o} = o$, and train using REINFORCE \citep{williams1992simple}. The agents co-ordinate to form a language $\mathcal{G}$ comprising pairs of objects and messages $\mathcal{G} = \{ (o_1, m_1), \dots, (o_N, m_N) \}$, where $N = n_{val}^{n_{att}}$ is the number of objects in the object space $\mathcal{O}$ \citep{lazaridou2018_refgames}.




In our work, we will consider the Sender or Receiver models in isolation, and attempt to obtain insights into their intrinsic compositional inductive biases.

\subsection{Compositionality metrics} \label{section:compositionality}





To measure compositionality, \citet{andreas_measuring_compositionality_2019} proposed TRE. TRE is a mathematical implementation of a definition of compositionality that the whole is a composition of the parts. TRE imposes no constraints on the composition function. Practical implementations of TRE provide opinions on allowed composition function. Section 7 of \citet{andreas_measuring_compositionality_2019} (hereafter `TRE7') takes the composition function to be the concatenation of sub-messages, followed by parameterized permutation. 
\citet{chaabouni2020compositionality}'s posdis assumes a message whose length equals the number of attributes in the input object, and where each message token, in a specific position, represents a single attribute.
Their bosdis constrains the meaning of a token to be invariant with position.
Thus, these metrics assume that we can partition messages into groups of one or more message tokens that each represent one attribute. \citet{resnick2019} indeed explicitly incorporate such a partition function into their resent metric. Lastly, \citet{brighton_kirby_2006_topographic_mappings} proposed topsim (`topological similarity'), which is a mature, widely-used metric, with few assumptions. topsim reports the correlation between the distances between objects, and distances between messages, over pairs of (object, message) tuples. The distance between messages is typically taken to be the L1 norm, or an edit distance. topsim will be a maximum when groups of message tokens map to individual attributes, and are combined with concatenation, possibly followed by permutation, similar to TRE7. All assume a permutation over concatenation as the composition function.


We will see in our experiments that it is possible to apply simple transforms to messages, which do not affect much the acquisition speed of neural models. However, which render the message apparently non-compositional to humans, and to our current metrics of compositionality.


\subsection{Other work on compositionality}

One approach to investigating the compositional inductive biases of models is to run many emergent communication experiments. This is time-consuming, noisy, and entangles many factors of variation.
Importantly, it is unclear how to inspect the compositional characteristics of the resulting languages.
We choose an alternative approach of generating languages which exhibit specific deviations from a perfectly compositional language; and measuring how easily each model can fit these artificial languages.
Our approach is similar to that used in \citet{libowling_easeofteaching}, \citet{kharitonov-baroni-2020-acquisition} and \citet{resnick2019}. However, \citet{libowling_easeofteaching} only considers a single transformation (permutation); \citet{kharitonov-baroni-2020-acquisition} focus on the effect of compositionality on generalization; and \citet{resnick2019} investigates primarily the effect of capacity.
\citet{hupke_compositionality_decomposed_2020} and \citet{white_inductive_bias_neural_lm_artificial_lang} use artificially created languages to test neural model's understanding of compositional forms that appear in natural language.
In our work, we search for languages which models can fit to easily, but which a human might consider non-compositional.

\section{Methodology}

In our work, we will train the Sender or Receiver in isolation, using artificial languages of our choosing.
We seek grammars which score poorly on compositional metrics, appear non-compositional to humans, but which demonstrate a fast acquisition speed by neural networks.
The general approach we follow is to start with concatenation grammars, and apply transformations to the linguistic representations which we hope might not affect the compositional form, as perceived by neural networks.

\subsection{Artificial grammars}

\subsubsection{Concatenation grammar (\textsc{concat})}

We start from a simple concatenation composition.
We sample a bijective map from each $\Sigma_{i,j}$ to sub-messages $w_{i,j}$, of length $c_w$, drawn from vocabulary $V$, where $c_w = c_{len} / n_{att}$.
Given an object $o$, we map each attribute value $o^{(j)}$ to a sub-message $w_{j, o^{(j)}}$ (i.e. the word for attribute $j$ and attribute value $o^{(j)}$), and concatenate the sub-messages. For example, attribute value `red' could map to sub-sequence `adaa', and `box' could map to sub-message `ccad'. Thus object (red, box) would map to message `adaaccad', and any red object would have a message starting with `adaa'.

\subsubsection{Holistic grammar (\textsc{hol})}

For each object $o_n$ we generate a random message $m_n$. This provides a baseline to compare the acquisition speed on other grammars against.

\subsubsection{Permuted grammar (\textsc{perm})}

We sample a single permutation, and apply this to all messages in a sampled concatenation grammar $\mathcal{G}_{concat}$, to form a permuted language $\mathcal{G}_{perm}$.

\subsubsection{Random Projection grammar (\textsc{proj})}

Neural networks apply projections at each layer. We hypothesize therefore that the ground truth output given to a neural network can be arbitrarily projected, without affecting the acquisition speed. Let use first consider the general, non-discrete, case, given dataset $\mathcal{D} = \{(\x_1, \y_1), \dots, (\x_N, \y_N)\}$, where each $(\x_n, \y_n)$ is a pair of input and output vectors. We hypothesize that we can apply any non-singular projection matrix $P$ to all $y_n$, forming $y'_n = Py_n$, without affecting the acquisition speed of a neural network.

In the case of a discrete message $m_n$,
we first expand to one-hot, vectorize, then apply. We form a one-hot matrix $m_n \rightarrow m_n^{onehot} \in \R^{c_{len} \times |V|}$, adding a new dimension over $V$. We vectorize to form $\vec(m_n^{onehot}) \in \R^{(c_{len} |V|)}$, then apply a projection $P \in \R^{(c_{len} |V|) \times (c_{len} |V|)}$. After unvectorizing and taking the argmax to recover a new discrete message, we obtain:

$$
m_n^{proj} = \arg \max_{V} \vec^{-1}(P \vec(m_n^{onehot}))
$$

We sample a single projection matrix $P$ per generated language. To the best of our knowledge, there is no equivalent composition operator to \textsc{proj} in natural language.

\subsubsection{Cumulative Rotation grammar (\textsc{rot})}


Vanilla recurrent neural networks (RNNs) take the input from the previous time-step and project it. Consider a transformation where we add the transformed output from the previous timestep to the current timestep:

$$
m_n^{(j,rot)} = (m_n^{(j - 1, rot)} + m_n^{(j)}) \mod |V|
$$

(where $m^{(j)}$ is the message symbol at position $j$, and $m^{(j,rot)}$ is the message symbol at position $j$ in the cumulatively rotated message. $\mod$ is the modulo operator).

We hypothesize that such a transformation is aligned with the transformations in a vanilla RNN, and so might be acquired quickly.
Meanwhile, \textsc{rot} has no equivalent composition function in human natural language.





\subsubsection{Relocatable atomic groups of tokens (\textsc{shufdet})}

We would like to encourage the models to emerge relocatable atomic groups of tokens, that is something similar to words in natural language. We want a deterministic shuffling, so that the Sender model knows which variation to output. In natural language, some word orders are dependent on the values of certain words. For example, in French, the adjective `neuve' follows a noun, whereas `nouvelle' precedes it. Thus we use the value of the last attribute of the meaning to determine the order of the sub-messages $w$, prior to concatenation. That is, for each possible value of the last attribute, we sample a permutation, and we apply this same permutation to all messages having the same last attribute value.

\textsc{shufdet} contrasts with the other artificial grammars we propose in that we feel that models with a similar compositional inductive bias to humans should acquire these grammars quickly. In Appendix \ref{app:shuf} we present an additional variation \textsc{shuf}.

\subsection{Compositionality metrics}

In addition to measuring model acquisition speed, we evaluate samples of each of the grammars for compositional metrics: bosdis, posdis, TRE7 and topsim.
Since we have $c_{len} > n_{att}$, we violate assumptions of bosdis and posdis.
However, we provide their scores for completeness.
We wanted to use in addition resent.
However, minimizing over all possible message partitions took combinatorial time.
Therefore we relaxed the minimization, to give a new metric HCE, which we describe in Appendix \ref{app:hce}.

\subsection{Neural models under test}

We primarily target neural models frequently used in emergent communication and natural language processing.
In addition we experiment with the evolved Sender model from \citet{dagan2020co}, an RNN decoder with zero'd inputs, and a novel architecture, HU-RNN.



\subsubsection{RNN decoder with zero'd inputs (RNNZero)}

An RNN comprises an inner cell $o_t, h_t = \rnn(x_t, h_{t-1})$, where $o_t$ is output at time step $t$, $h_t$ is hidden state, and $x_t$ is input. When used as a decoder, the output is fed back auto-regressively: $o_t, h_t = \rnn(W_{hi} o_{t-1}, h_{t-1})$, where $W_{hi}$ is a projection. We experiment in addition with a decoder where the input at each time step is all zeros: $o_t, h_t = \rnn(\mathbf{0}, h_{t-1})$. We use a `-Z' suffix to denote this, e.g. `LSTM-Z', when using an LSTM-based decoder \citep{hochreiter1997long}.

In many frameworks, e.g. PyTorch \citep{pytorch}, RNN-Zs uses fewer lines of code, and arguably have lower Kolmogorov complexity \citep{kolmogorov1963tables}. We show that their compositional inductive bias is sometimes better than the auto-regressive variant.

\subsubsection{Hierarchical Unit RNN (\textsc{HU-RNN})}

Hierarchical-Unit RNNS (`HU-RNNs') are fully differentiable, and might encourage an inductive bias towards receiving and sending atomic relocatable groups of tokens, i.e. for \textsc{shufdet}.

`HUSendZ' is a Sender model. There are two recurrent neural network layers (`RNN') \citep{hopfield1982neural} layers. Conceptually, the lower layer, $\rnn_l$, decodes word embeddings, and the upper layer, $\rnn_u$, decodes tokens. A scalar `stopness', $s_t$, gates the feed of the word embedding from the lower to the upper layer. $s_t$ is generated by the upper layer. The lower hidden state is initialized from an input embedding, and the upper state is initialized as all zeros. At each time step:

\begin{align}
h^{(l)}_t &= (1 - s_{t-1} )\cdot h^{(l)}_{t-1} + s_{t-1} \cdot \rnn_u(\mathbf{0}, h^{(l)}_{t-1}) \nonumber \\
h^{(u)}_t &= \rnn_l\left(\mathbf{0}, (1 - s_{t-1}) \cdot h^{(u)}_{t-1} + s_{t - 1} \cdot h^{(l)}_t\right) \nonumber \\
s_t &= \sigma(f_s(h^{(u)}_t)) \nonumber\qquad
 \hat{y}_t = o(h^{(u)}_t) \nonumber
\end{align}

where $o(\cdot)$ and $f_h(\cdot)$ are projection layers. HUSendA is an auto-regressive variant of HUSendZ, in which the input to $\rnn_l$ at each timestep is a projection of $\hat{y}_{t-1}$, instead of $\mathbf{0}$. Figure \ref{fig:hurnn_sender} depicts the HU-RNN Sender architecture graphically.

Note that we can choose any RNN for $\rnn_u$ and $\rnn_l$. We use the suffix `:[rnn type]', where `:RNN' means a vanilla RNN, `:LSTM' is an LSTM, and `:dgsend' means using the Sender RNN from \citet{dagan2020co}.

We also propose HURecv, which is a Receiver model, see Appendix \ref{app:hu_receiver}.

\section{Experiments}

Code for experiments is at \footnote{https://github.com/asappresearch/compositional-inductive-bias}.

\subsection{Examples of grammars}

\begin{table}
\small
\caption{Example utterances from an instance of each grammar, for 4 different objects. Best viewed in color.}
\centering
\begin{tabular}{llllllll}
\toprule 
\textsc{Object} & \textsc{concat} & \textsc{perm} & \textsc{rot} & \textsc{proj} & \textsc{shufdet} & \textsc{hol} \\ 
\midrule 
(\teal{0}, \purple{0}, \orange{0}) & \teal{dad}\purple{acb}\orange{bba} &
                            \orange{a}\purple{a}\orange{b}\teal{d}\purple{c}\teal{d}\teal{a}\orange{b}\purple{b} & ddccabcdd & dcbcbbaad & 
                            \teal{dad}\purple{acb}\orange{bba} & adbcddadc \\ 
(\teal{0}, \purple{0}, \blue{1}) & \teal{dad}\purple{acb}\blue{cca} &
                            \blue{a}\purple{a}\blue{c}\teal{d}\purple{c}\teal{d}\teal{a}\blue{c}\purple{b} & ddccabdbb & bdbcabaad & 
                            \blue{cca}\teal{dad}\purple{acb} & bcaadacba \\ 
(\teal{0}, 1, \orange{0}) & \teal{dad}cab\orange{bba} &
                            \orange{a}c\orange{b}\teal{d}a\teal{d}\teal{a}\orange{b}b & ddcaabcdd & dbbcabcad &
                            \teal{dad}cab\orange{bba} & bcaccaddb \\ 
(1, \purple{0}, \orange{0}) &ddb\purple{acb}\orange{bba} &
                            \orange{a}\purple{a}\orange{b}d\purple{c}bd\orange{b}\purple{b} & dcddbcdaa & acbcabaad &
                            ddb\purple{acb}\orange{bba} & daaaacbdc \\ 
\bottomrule 
\end{tabular}
\label{tab:grammar_examples_3x10}
\end{table}


Table \ref{tab:grammar_examples_3x10} shows examples of each grammar, for 4 objects.
For \textsc{concat}, changing one attribute changes 3 adjacent utterance tokens. \textsc{perm} rearranges columns of \textsc{concat} utterance tokens. \textsc{shufdet} rearranges blocks of 3 utterance tokens, as a function of the last object attribute. We depict utterances for $n_{att}=3$, and $c_{len} = 3 \cdot n_{att}$. In our experiments we use $n_{att}=5$ and $c_{len}=4 \cdot n_{att}$. Examples for this geometry can be found in Appendix \ref{app:example_utterances}.

\subsection{Compositional metric evaluation}

%

\begin{figure}[ht]
\tiny
\centering
\begin{tikzpicture}
\begin{axis}[
     axis y line*=left,
    ybar = 0cm,
    ylabel=Compositionality,
    enlarge y limits  = 0.0,
    ymax = 1.1,
    ymin=-0.03,
    bar width=0.15cm,
    xtick=data,
    legend style={at={(2,0)},anchor=north west},
    symbolic x coords={
        \textsc{concat},
        \textsc{perm},
        \textsc{rot},
        \textsc{proj},
        \textsc{shufdet},
        \textsc{hol},
    },
    width=0.5\textwidth,
    height=0.3\textwidth
]
\pgfplotsset{
    legend image code/.code={
        \draw [#1] (0cm,-0.1cm) rectangle (0.3cm,0.025cm);
    },
}
  \addplot [color=red, pattern color=.,
    postaction={
        pattern=dots,
    }] table[x=grammar, y=posdis,col sep=comma]{data/metrics_5x10.csv}; \label{p1};
  \addplot [color=blue, pattern color=.,pattern=horizontal lines]  table[x=grammar, y=bosdis,col sep=comma]{data/metrics_5x10.csv}; \label{p2};
  \addplot [pattern=none] table[x=grammar, y=compent,col sep=comma]{data/metrics_5x10.csv}; \label{p3};
  \addplot table[x=grammar, y=topsim,col sep=comma]{data/metrics_5x10.csv}; \label{p4};

  \end{axis}
  
\begin{axis}[
    axis y line*=right,
    axis x line=none,
    ylabel=\textsc{tre7},
    ybar = 0cm,
    bar shift=1.26cm,
    ymin=-0.7,
    ymax = 25,
    enlarge y limits  = 0.0,
    legend style={draw=none},
    bar width=0.15cm,
    xtick=data,
    legend style={at={(1.18,0.8)},anchor=north west},
    symbolic x coords={
        \textsc{concat},
        \textsc{perm},
        \textsc{rot},
        \textsc{proj},
        \textsc{shufdet},
        \textsc{hol},
    },
    width=0.5\textwidth,
    height=0.3\textwidth
  ]
\pgfplotsset{
    legend image code/.code={
        \draw [#1] (0cm,-0.1cm) rectangle (0.4cm,0.1cm);
    },
}

\addlegendimage{/pgfplots/refstyle=p1}\addlegendentry{\textsc{posdis}}
\addlegendimage{/pgfplots/refstyle=p2}\addlegendentry{\textsc{bosdis}}
\addlegendimage{/pgfplots/refstyle=p3}\addlegendentry{\textsc{HCE}}
\addlegendimage{/pgfplots/refstyle=p4}\addlegendentry{\textsc{topsim}}

\addplot [fill=green!10] table[x=grammar, y=tre7,col sep=comma]{data/metrics_5x10.csv}; \label{p5};

\addlegendimage{/pgfplots/refstyle=p5}\addlegendentry{\textsc{tre7}}

\end{axis}
 
\end{tikzpicture}
\caption{Values of compositionality metrics for each of the artificial grammars. Each bar is an average over 5 seeds. \textsc{tre7} uses right-hand y-axis. High \textsc{tre7} means low compositionality}
\label{fig:metrics_5x10}
\end{figure}
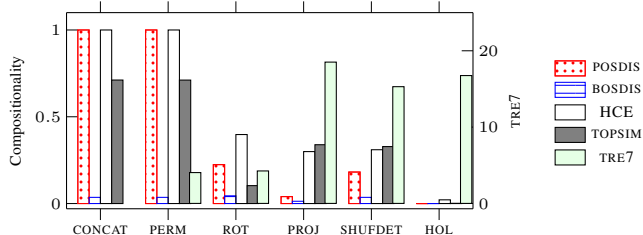

Figure \ref{fig:metrics_5x10} shows the values of compositional metrics for samples from our artificial grammars, using $n_{att}=5$, $n_{val}=10$.
The compositionality metrics show low compositionality for all the artificial grammars, except for \textsc{concat} and \textsc{perm}. Thus our transformations successfully hide the compositional structure from current compositional metrics.

\subsection{Human evaluation}

\begin{figure}
    \centering
    \includegraphics[width=0.45\columnwidth]{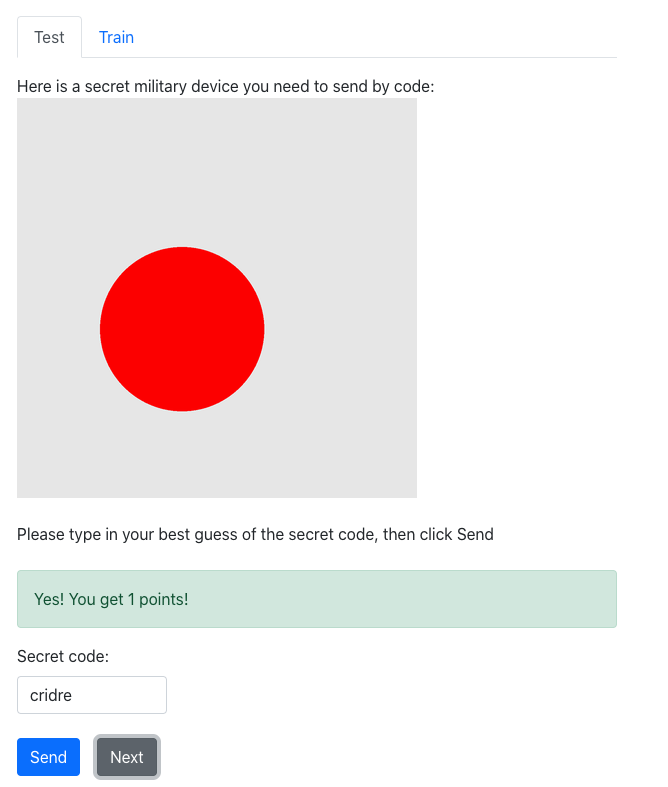}
    \caption{User interface for our `Secret Spy Codes' game. This image depicts a \textsc{perm} grammar, where the original utterance was `redcir'.}
    \label{fig:mturk}
\end{figure}

We constructed an MTurk \citep{mturk} task, `Secret Spy Codes', in order to evaluate human performance on \textsc{Icy} grammars. Figure \ref{fig:mturk} shows the user interface. Human subjects were tasked with writing out the code that represents depicted geometric objects.
They needed substantial effort in order to learn just a few new words.
Thus, we used objects with only two attributes: shape and color; and we experimented with using abbreviated English words, which were easier to learn.

\subsubsection{Dataset}

\textsc{synth} uses artificially generated random words for each attribute value. We sample 2-letter words from a vocabulary size of 4. Each utterance therefore has 4 letters: 2 for shape, and 2 for color. Since humans found these words challenging to learn, so we used just three possible values for each attribute, giving 9 combinations in total.

\textsc{eng} uses 3-letter English abbreviations for attribute values, e.g. `tri' for `triangle', and `grn' for `green'. The words describing each attribute value in \textsc{eng} are relatively easy for a human to learn. Therefore, we used 5 attribute values for each attribute, giving 25 possible combinations.

We held out three color-shape combinations, that were not made available during training. For example, subjects might have access to a red circle and a blue triangle, but not a red triangle. Subjects who could perceive the compositional structure of a grammar should be able to get these holdout instances correct.

\subsubsection{Evaluation}


\begin{table*}[htb!]
\small
\centering
\begin{tabular}{ llllll }
\toprule
Dataset & \textsc{concat} & \textsc{perm} & \textsc{proj} & \textsc{rot} & \textsc{shufdet} \\ 
\midrule 
\textsc{synth} & $\mathbf{0.2\pm0.1}$ & $0.04\pm0.05$ & $0.02\pm0.04$ & $0.04\pm0.07$ & $0.06\pm0.06$ \\ 
\textsc{eng} & $\mathbf{0.6\pm0.2}$ & $0.2\pm0.2$ & $0.04\pm0.08$ & $0.02\pm0.03$ & $\mathbf{0.7\pm0.2}$ \\ 
\bottomrule
\end{tabular}
\caption{Human evaluation results. Values are $\acc_{holdout}$}
\label{tab:turk_results}
\end{table*}

We measured subjects' accuracy on the 3 held out examples. The results are shown in Table \ref{tab:turk_results}. For \textsc{synth}, $\acc_{holdout}$ was poor for all grammars: humans were unable to spot compositional form using unfamiliar words. In \textsc{eng}, $\acc_{holdout}$ was high for both \textsc{concat} and \textsc{shufdet} grammars, as expected, and low for all other grammars. This shows that the composition functions in \textsc{perm}, \textsc{proj} and \textsc{rot} were not clearly apparent to human subjects, even though, as we shall see next, neural models can acquire these grammars easily.




\subsection{Neural model evaluation}


\begin{table}
\small
\caption{Compositional inductive bias for some standard Sender models, for $n_{att}=5$, $n_{val}=10$. Results are each averaged over 10 seeds. Results are the ratio of the convergence time for the grammar relative to \textsc{concat}, therefore have no units. CI95 is in Table \ref{tab:sender_5x10_ci95}.}
\begin{tabular}{p{0.4\textwidth}rrrrrll}
\textbf{Model}                                           & \multicolumn{1}{l}{\textbf{Params}} & \multicolumn{1}{l}{\textbf{\textsc{perm}}} & \multicolumn{1}{l}{\textbf{\textsc{proj}}} & \multicolumn{1}{l}{\textbf{\textsc{rot}}}        & \multicolumn{1}{l}{\textbf{\textsc{shufdet}}}    & \textbf{\textsc{hol}}        &  \\
1-layer MLP \citep{rumelhart1986learning}                & 5100                                & \cellcolor[HTML]{D9EAD3}1.12               & \cellcolor[HTML]{D9EAD3}1.9                & \multicolumn{1}{l}{\cellcolor[HTML]{F4CCCC}> 20} & \multicolumn{1}{l}{\cellcolor[HTML]{F4CCCC}> 20} & \cellcolor[HTML]{F4CCCC}> 20 &  \\
2-layer MLP                                              & 19428                               & \cellcolor[HTML]{D9EAD3}1                  & \cellcolor[HTML]{D9EAD3}2.1                & \multicolumn{1}{l}{\cellcolor[HTML]{F4CCCC}> 20} & 7                                                & \cellcolor[HTML]{F4CCCC}> 20 &  \\
1-layer LSTM \citep{hochreiter1997long}                  & 139909                              & \cellcolor[HTML]{D9EAD3}1                  & \cellcolor[HTML]{D9EAD3}2.2                & 7                                                & \cellcolor[HTML]{D9EAD3}1.6                      & \cellcolor[HTML]{F4CCCC}> 20 &  \\
2-layer LSTM                                             & 272005                              & \cellcolor[HTML]{D9EAD3}0.9                & \cellcolor[HTML]{D9EAD3}1.9                & \cellcolor[HTML]{D9EAD3}4.6                      & \cellcolor[HTML]{D9EAD3}1.36                     & \cellcolor[HTML]{F4CCCC}> 20 &  \\
1-layer Transformer decoder \citep{vaswani2017attention} & 272389                              & \cellcolor[HTML]{D9EAD3}1.02               & \cellcolor[HTML]{D9EAD3}2.5                & 10                                               & \cellcolor[HTML]{D9EAD3}2.1                      & \cellcolor[HTML]{F4CCCC}> 20 &  \\
2-layer Transformer decoder                              & 536965                              & \cellcolor[HTML]{D9EAD3}1.08               & \cellcolor[HTML]{D9EAD3}2.2                & 10.4                                             & \cellcolor[HTML]{D9EAD3}1.98                     & \cellcolor[HTML]{F4CCCC}> 20 &  \\
Hashtable                                                & O(N)                                   & 1                                          & 0.98                                       & 0.98                                             & 1.02                                             & \multicolumn{1}{r}{1.06}     & 
\end{tabular}
\label{tab:sender_5x10}
\end{table}

We use the \sys/ benchmark to evaluate standard neural models for specific aspects of their compositional inductive bias.
We focus on Sender models in our presentation.
Results for Receiver models are in Appendix \ref{app:additional_results}.
We train each model supervised on a specific artificial grammar from \sys/, using cross-entropy loss.

We count the number of training steps, $N_{acquire}$, required to train each grammar to a training accuracy of $\acc_{tgt}$, where accuracy is token-level accuracy. For each grammar, $\mathcal{G}$, we report the ratio
$b^{(\mathcal{G})} = N_{acquire}^{(\mathcal{G})} / N_{acquire}^{(\mathcal{G}_{\textsc{concat}})}$.
We used $n_{att} = 5$ and $n_{val} = 10$, $c_{len} = 20$, $V = 4$, and $\acc_{tgt} = 0.8$. We halt training if $b^{(\mathcal{G})}$ reaches $20$.

Table \ref{tab:sender_5x10} shows the results.
Detailed architectural descriptions of the `Model' column are provided in Appendix \ref{app:sender_model_architectures}.
The remaining columns, except for `Params', show the acquisition time, $b$, for each  grammar, relative to \textsc{concat}. We have highlighted in red the scenarios that failed to reach convergence; and in green the scenarios where $b$ was less than 1/3 that of \textsc{hol}, which shows that language acquisition was relatively fast.

We can see that for many models, our transformations do not much affect the acquisition speed by neural networks. Therefore, in an emergent communication scenario, neural models can generate languages which appear non-compositional both to our current metrics, and to human evaluation. Such languages will therefore be deemed `non-compositional' by all current evaluation methods, except for generalization. This might explain the empirically observed lack of correlation between measured language compositionality, and generalization, in emergent communication experiments.

\subsection{Results are independent of number of parameters}

\begin{table}
\small
\caption{Effect of number of parameter on compositional inductive bias. Results are each averaged over 5 seeds. CI95 is in Table \ref{tab:params_change_ci95}}.
\begin{tabular}{lrrrrlrl}
\textbf{Model} & \multicolumn{1}{l}{\textbf{Emb size}} & \multicolumn{1}{l}{\textbf{Params}} & \multicolumn{1}{l}{\textbf{\textsc{perm}}} & \multicolumn{1}{l}{\textbf{\textsc{proj}}} & \textbf{\textsc{rot}}                           & \multicolumn{1}{l}{\textbf{\textsc{shufdet}}} & \textbf{\textsc{hol}}        \\
1-layer LSTM   & 128                                   & 139909                              & \cellcolor[HTML]{D9EAD3}1                  & \cellcolor[HTML]{D9EAD3}2.2                & \multicolumn{1}{r}{7}                           & \cellcolor[HTML]{D9EAD3}1.6                   & \cellcolor[HTML]{F4CCCC}> 20 \\
1-layer LSTM   & 1280                                  & 13195525                            & \cellcolor[HTML]{D9EAD3}0.94               & \cellcolor[HTML]{D9EAD3}1.6                & \multicolumn{1}{r}{\cellcolor[HTML]{D9EAD3}2.7} & \cellcolor[HTML]{D9EAD3}1.4                   & \cellcolor[HTML]{F4CCCC}> 20 \\
2-layer MLP    & 128                                   & 19428                               & \cellcolor[HTML]{D9EAD3}1                  & \cellcolor[HTML]{D9EAD3}2.1                & \cellcolor[HTML]{F4CCCC}> 20                    & 7                                             & \cellcolor[HTML]{F4CCCC}> 20 \\
2-layer MLP    & 1280                                  & 193380                              & \cellcolor[HTML]{D9EAD3}1.02               & \cellcolor[HTML]{D9EAD3}1.86               & \cellcolor[HTML]{F4CCCC}> 20                    & 10                                            & \cellcolor[HTML]{F4CCCC}> 20
\end{tabular}
\label{tab:num_parameters}
\end{table}

An obvious concern with Table \ref{tab:sender_5x10} is that the number of parameters varies between models, so we vary the parameters, by changing the hidden size. Table \ref{tab:num_parameters} shows the results.
We can see that the relative acquisition speed, relative to \textsc{concat}, is not changed much by a 10-fold increase in parameters,
relative to the differences between the architectures. This is encouraging: we are not simply viewing an artifact of model size.

\subsection{RNNZero increases bias against \textsc{perm}}

\begin{table}
\small
\caption{Zero-RNN improves bias against \textsc{perm}. Results are mean over 5 runs. CI95 Table \ref{tab:zerornn_ci95}.}
\begin{tabular}{lrrrrrl}
\textbf{Model}  & \multicolumn{1}{l}{\textbf{Parameters}} & \multicolumn{1}{l}{\textbf{\textsc{perm}}} & \multicolumn{1}{l}{\textbf{\textsc{proj}}} & \multicolumn{1}{l}{\textbf{\textsc{rot}}} & \multicolumn{1}{l}{\textbf{\textsc{shufdet}}} & \textbf{\textsc{hol}}       \\
RNN             & 40837                                   & \cellcolor[HTML]{D9EAD3}0.76               & \cellcolor[HTML]{D9EAD3}2.8                & 18                                        & \cellcolor[HTML]{D9EAD3}1.74                  & \cellcolor[HTML]{F4CCCC}>20 \\
RNN-Z           & 40069                                   & \cellcolor[HTML]{D9EAD3}0.8                & \cellcolor[HTML]{D9EAD3}2.9                & 19                                        & \cellcolor[HTML]{D9EAD3}2                     & \cellcolor[HTML]{F4CCCC}>20 \\
GRU \citep{gru} & 106885                                  & \cellcolor[HTML]{D9EAD3}1                  & \cellcolor[HTML]{D9EAD3}2.4                & \cellcolor[HTML]{D9EAD3}6                 & \cellcolor[HTML]{D9EAD3}1.88                  & \cellcolor[HTML]{F4CCCC}>20 \\
GRU-Z           & 106117                                  & \cellcolor[HTML]{D9EAD3}1.16               & \cellcolor[HTML]{D9EAD3}2.3                & \cellcolor[HTML]{D9EAD3}4.7               & \cellcolor[HTML]{D9EAD3}1.86                  & \cellcolor[HTML]{F4CCCC}>20 \\
LSTM            & 139909                                  & \cellcolor[HTML]{D9EAD3}1                  & \cellcolor[HTML]{D9EAD3}2.2                & 7                                         & \cellcolor[HTML]{D9EAD3}1.6                   & \cellcolor[HTML]{F4CCCC}>20 \\
LSTM-Z          & 139141                                  & \cellcolor[HTML]{D9EAD3}1.18               & \cellcolor[HTML]{D9EAD3}2.3                & 7                                         & \cellcolor[HTML]{D9EAD3}1.82                  & \cellcolor[HTML]{F4CCCC}>20
\end{tabular}
    \label{tab:rnnz_perm}
\end{table}

Table \ref{tab:rnnz_perm} shows the effect of not feeding the output of an RNN decoder as the input at each step. Surprisingly this increases bias against \textsc{perm}. That is, actually increases a prior over adjacency.

\subsection{HUSendZ:dgsend has low bias against \textsc{shufdet}}

\begin{table}
\small
\caption{Selected results for low \textsc{shufdet} bias, mean over 10 runs. Full results Table \ref{tab:shufdet_ci95}.}
\begin{tabular}{lrrrrrl}
\textbf{Model}     & \multicolumn{1}{l}{\textbf{Params}} & \multicolumn{1}{l}{\textbf{\textsc{perm}}} & \multicolumn{1}{l}{\textbf{\textsc{proj}}} & \multicolumn{1}{l}{\textbf{\textsc{rot}}} & \multicolumn{1}{l}{\textbf{\textsc{shufdet}}} & \textbf{\textsc{hol}}        \\
LSTM    & 139909                     & \cellcolor[HTML]{D9EAD3}1.08          & \cellcolor[HTML]{D9EAD3}2.4          & \cellcolor[HTML]{D9EAD3}6.3      & \cellcolor[HTML]{D9EAD3}1.8           & \cellcolor[HTML]{F4CCCC}> 20 \\
dgsend  & 106117                     & \cellcolor[HTML]{D9EAD3}0.92          & \cellcolor[HTML]{D9EAD3}2.3          & \cellcolor[HTML]{D9EAD3}5.5      & \cellcolor[HTML]{D9EAD3}1.52          & \cellcolor[HTML]{F4CCCC}> 20 \\
HUSendZ:RNN       & 40710                      & \cellcolor[HTML]{D9EAD3}0.93          & \cellcolor[HTML]{D9EAD3}\textbf{2.5} & \textbf{15}                      & \cellcolor[HTML]{D9EAD3}1.68          & \cellcolor[HTML]{F4CCCC}> 20 \\
HUSendZ:dgsend    & 138758                     & \cellcolor[HTML]{D9EAD3}1.03          & \cellcolor[HTML]{D9EAD3}1.67         & \cellcolor[HTML]{D9EAD3}4.5      & \cellcolor[HTML]{D9EAD3}\textbf{1.27} & \cellcolor[HTML]{F4CCCC}> 20
\end{tabular}
\label{tab:shufdet_results}
\end{table}

We searched for neural models with reduced bias against \textsc{shufdet}, including using RNN-Z, dgsend \citep{dagan2020co}, and HU-RNN. Table \ref{tab:shufdet_results} shows a sub-set of the results. More results are in Appendix \ref{app:shuf}. 
`dgsend' acquired \textsc{shufdet} faster than LSTM.
HUSend using a vanilla RNN as $\rnn_l$ and $\rnn_u$ acquired \textsc{shufdet} faster than LSTM.
Combining HUSendZ with dgsend acquired \textsc{shufdet} fastest.

\subsection{End-to-end training}

We experimented with measuring the compositional inductive bias of a Sender and Receiver model placed end to end, see Appendix \ref{app:endtoend}

\section{Conclusion}

We have shown that it is possible to construct transformations that, when applied to concatenation grammars, result in grammars that machines can learn easily but which humans find challenging to learn. This could explain the disconnect highlighted in recent papers between neural network ability to generalize, in an emergent communication context, and the compositionality of the resulting languages, as measured by recent metrics of compositionality. We propose to use the families of transformations as a benchmark, \textsc{Icy}, for measuring aspects of the compositional inductive bias of neural networks, and searching for models with similar biases to humans. We use our benchmark to propose one such neural model, \textsc{hu-rnn}, which shows a compositional inductive bias towards relocatable atomic word-like groups of tokens.





\section{Reproducibility}

Full code is provided in the addendum, along with instructions in the README.md. Full code will be published to github following acceptance. Each experiment was run multiple times (usually 5 or 10), using different seeds, and the mean reported. CI95 ranges are available in Appendix \ref{app:ci95}.

\section{Ethics}

This work does involve human subjects, who needed to learn to use artificially generated codes to label abstract geometric objects. The annotation device was created as a game, that many people found fun to play. We received many feedbacks stating `good', `very interesting task'. None of the language or figures being trained on contain any obvious characteristics which could be deemed racist, sexist, or having any other obvious human-centric harmful biases, as far as we can tell.


This work contains no obviously harmful insights, methodologies or applications. There are no obvious conflicts of interest or sponsorship to note. There are no obvious discrimination/bias/fairness concerns to report. There are no obvious issues with privacy, security, or legal compliance. All data provided was artificially generated, and does not present privacy or other issues. We have done our due diligence to ensure the integrity and reproducibility of our research.

Although emergent communication investigates the communications between neural models, who learn to generate new languages, as part of collaborative tasks, we do not believe that such models are `alive', or `conscious', though we admit that we do not have any way to determine this in any objective way. The number of neurons of the models concerned was orders of magnitude less than that of the human brain. The models were not exposed to sufficiently varied or complex data that we feel that they could have learned advanced sentience or perception, although again we admit that we are not aware of an objective `threshold' or similar that we could compare with.

\bibliography{bias}
\bibliographystyle{iclr2022_conference}

\clearpage

\appendix

\section{hyperparameters} \label{app:hyperparams}

\subsection{General hyper-parameters}

\begin{table}[htb!]
\small
    \caption{General hyper-parameters, for sender and receiver network experiments}
\centering
    \begin{tabular}{ll}
    \toprule
        Setting & Value \\
        \midrule
        Embedding size, $d_{emb}$ & 128 \\
        Vocab size, $V$ & 4 \\
        Utterance length, $c_{len}$ & 4 * $n_{att}$ \\
        Dropout & 0 \\
        Gradient clipping & 5.0 \\
        Optimizer & Adam \\
        Batch size & 128 \\
    \bottomrule
    \end{tabular}
    \label{tab:general_hyperparameters} 
\end{table}

General hyper-parameters are shown in Table \ref{tab:general_hyperparameters}.

\section{Sender Model architectures} \label{app:sender_model_architectures}

We use a separate embedding matrix for each attribute, where the number of embeddings is equal to $n_{val}$. Given an object with $n_{att}$ attributes, we embed each of the attributes, then take the sum, to form a vector $e \in \R^{d_{emb}}$

\subsection{1-layer MLP}

Instead of embedding into $e \in \R^{d_{emb}}$, we embed into $\R^{c_{len} \cdot V }$, then we reshape into $\R^{c_{len} \times V}$.

\subsection{2-layer MLP}

We form $W^T \tanh(\text{drop}(e))$, where $W$ is a learnable matrix $\in \R^{d_{emb} \times (c_{len}\cdot V) }$. Then we reshape to be $\in \R^{c_{len} \times V }$

\subsection{1-layer LSTM}

We apply dropout to the embeddings $e$, then we use as the initial hidden state for the LSTM. At each timestep, we project the output token from the previous timestep (initially zero), and pass as the input token. We project the output at each timestep, to be in $\R^{V}$, and form the softmax, to obtain a probability distribution over tokens.


\subsection{2-layer LSTM}

2-layer version of the 1-layer LSTM above, where the output of the first layer at each timestep is fed into the input of the second layer. Each layer has its own hidden state and cell state.  We project the output from the second layer at each timestep, to be in $\R^{V}$, and form the softmax, to obtain a probability distribution over tokens.

\subsection{1- or 2-layer Transformer decoder}

TransDecSoft is a transformer decoder, as defined in \cite{vaswani2017attention}. Each softmaxed output token is passed in as input to the following timestep.

\subsection{hashtable}

Hashtable is a standard hashtable. We trained and scored using a similar approach to neural nets:
\begin{itemize}
    \item A minibatch of training examples was presented to the hashtable.
    \item The hashtable made a prediction. For previously unseen inputs, the hashtable predicted all 0s.
    \item The training accuracy was calculated using these predictions.
    \item The examples from this minibatch were added to the hashtable.
\end{itemize}

\subsection{Hierarchical-Unit RNN Sender, HU-Sender} \label{app:hu_sender}

\begin{figure}
    \centering
    \includegraphics[width=0.95\textwidth]{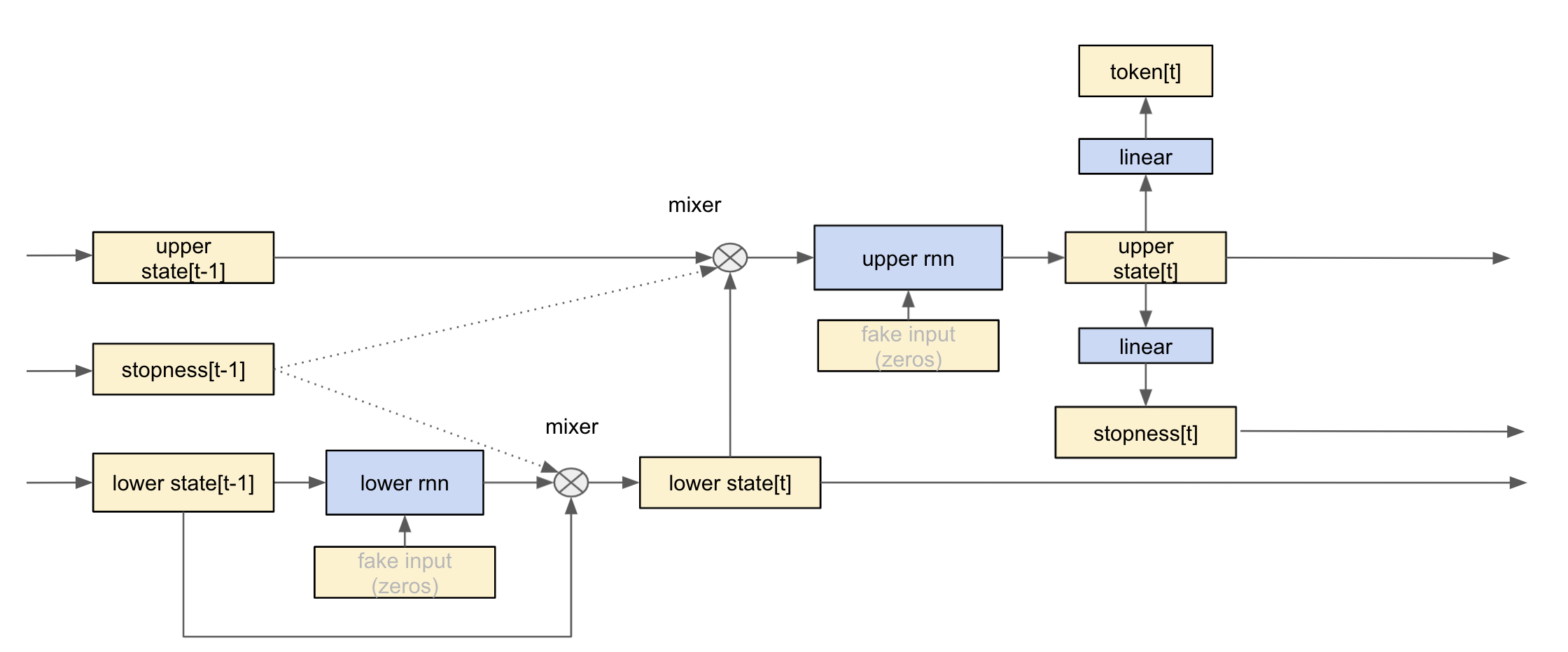}
    \caption{HU-RNN Sender Architecture}
    \label{fig:hurnn_sender}
\end{figure}

Figure \ref{fig:hurnn_sender} depicts the HU-RNN Sender architecture graphically.

\section{Receiver Model architectures} \label{app:receiver_model_architectures}

Given an input utterance of length $c_{len}$, vocab size $V$, in all cases, we first embed the tokens, to form a tensor $e \in \R^{c_{len} \times d_{emb}}$.

\subsection{CNN}

4 convolutional blocks, where each block consists of:
\begin{itemize}
    \item embed, as above
    \item 1d convolution (kernel size 3, padding 1, stride 1)
    \item max pooling (kernel size 2, padding 0, stride 2)
    \item ReLU activation
\end{itemize}

We only experiment with using a CNN as a receiver network.

\subsection{FC2L}

\begin{itemize}
    \item embed, as above, to form $e$
    \item form $\text{vec}(\tanh(\text{drop}(e)))$
    \item project, using learnable matrix $W$
    \item reshape to be $\in R^{n_{att} \times n_{val}}$
\end{itemize}

\subsection{RNNxL:rnntype}

Synonym for rnntype-xL, e.g. RNN2L:LSTM is equivalent to LSTM-1L. We first embed to form $e$ then pass the embedding for each timestep $t \in \{1, \dots, c_{len} \}$ into the RNN at each timestep. We take the final hidden state, apply dropout, and project using learnable matrix $W$ to be in $\R^{n_{att} \times n_{val}}$.

\subsection{SRU}

See \cite{sru}. An SRU is derived from an LSTM but with no connection between the hidden states. Connections between cell states remain. An SRU is much faster than an LSTM to train on a GPU. `RNN2L:SRU' is a synonym for a 2-layer SRU, i.e. SRU-2L. We treat SRU as any other RNN, see above.

\subsection{Hierarchical-Unit RNN Receiver, HU-Receiver} \label{app:hu_receiver}

HU-Sender was described in the main body.

HU-Receiver is a fully-differentiable hierarchical receiver. Incoming tokens, $x_t$ are fed in to a lower RNN, $\rnn_l$, one token per time step. The output state $h^{(l)}_t$ of $\rnn_l$ at each time step is used to generate a scalar `stopness', $s_t$, which conceptually represents the end of a word-like group of tokens. The upper state $h^{(u)}_t$ conceptually copies $h^{(u)}_{t-1}$ when $s_t$ is near $0$, or takes a step using an upper RNN, $\rnn_u$, when $s_t$ is near $1$. The formulae are thus:

\begin{align}
h^{(l)}_t &= \rnn_l(i(x_t), h^{(l)}_{t-1}) \nonumber \\
s_t &= \sigma(f_s(h^{(l)}_t)) \nonumber \\
\tilde{h}^{(u)}_t &= \rnn_u(h^{(l)}_t, h^{(u)}_{t-1}) \nonumber \\
h^{(u)}_t &= (1 - s_t) \cdot h^{(u)}_t + s_t \cdot \tilde{h}^{(u)}_t \nonumber
\end{align}

where $i(\cdot)$ and $f_s(\cdot)$ are projection layers, and $\sigma(\cdot)$ is the sigmoid function.

\begin{figure}
    \centering
    \includegraphics[width=0.95\textwidth]{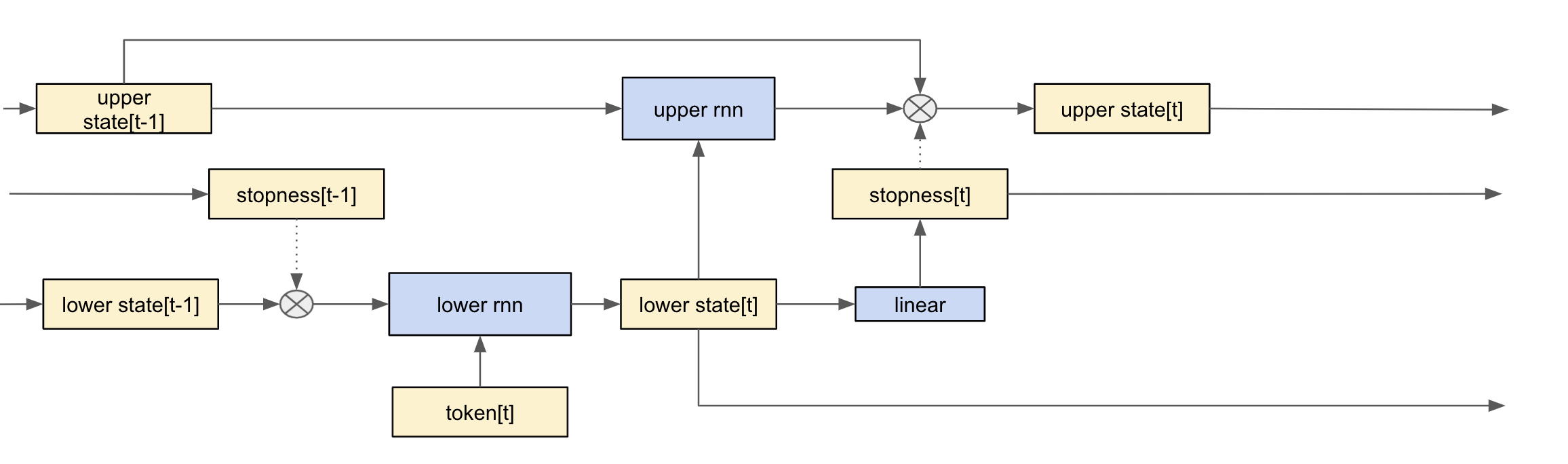}
    \caption{HU-RNN Receiver Architecture}
    \label{fig:hurnn_receiver}
\end{figure}

Figure \ref{fig:hurnn_receiver} depicts the HU-RNN Receiver model graphically.

\subsection{Hier:rnntype}

Synonym for the HU-Receiver model, described above. `rnntype' describes the RNN model used for each of $\rnn_l$ and $\rnn_u$.

\section{Human Compositional Entropy metric}\label{app:hce}

\subsection{Exposition}

We relax the minimization in the Residual Entropy metric \citep{resnick2019}. Resnick defines residual entropy as:

$$
re(\mathcal{M}, \mathcal{O}) = \min_{p \in \mathcal{P}} \frac{1}{n_{att}} \sum_{i=1}^{n_{att}} \frac{\mathcal{H}_{\mathcal{M}} (o^{(i)}|z[p_i])}{\mathcal{H}_{\mathcal{O}}(o^{(i)})}
$$

where $\mathcal{M}$ is the space of messages, $i$ is an index over attributes, $o^{(i)}$ is the i'th attribute, $\mathcal{P}$ is the set of all possible partitions over the messages, $p$ is one such partition, $p_i$ is one set in the partition $p$, $z[p_i]$ is the sub-sequence of each message indexed by set $p_i$, and $\mathcal{H}$ is the entropy. Thus, residual entropy finds a partition over messages, into $n_{att}$ sets, which associates each set in the partition with a specific attribute $\in \{1, \dots, n_{att}\}$, and minimizes the conditional entropy between each attribute in the dataset, and the corresponding message sub-sequences.

We can see that residual entropy assumes a composition that comprises a permutation over concatenation. This is thus a measure of human compositionality. It does not attempt to measure other members of the class of generalizable composition functions.

The minimization over $p \in \mathcal{P}$ is problematic because it involves a minimization over a combinatorial number of partitions. We seek to relax this, by using a greedy approach.

Similar to \citet{chaabouni2020compositionality} we form $\mathcal{I}(m^{(j)}; o^{(i)})$, the mutual information between the $j$'th symbol of each message, $m^{(j)}$, and the $i$'th attribute of each object, $o^{(i)}$, over the entire dataset:

$$
\mathcal{I}(m^{(j)}; o^{(i)}) = \sum_{n=1}^N p(m_n^{(j)}, o_n^{(i)}) \log
   \frac{p(m_n^{(j)}, o_n^{(i)})}
     {p(m_n^{(j)}) p(o_n^{(i)})}
$$

For each $m^{(j)}$, we calculate $o^{(j^*)} = \arg \max_{o^{(i)}} \mathcal{I}(m^{(j)}; o^{(i)})$. That is, $o^{(j^*)}$ is the attribute that has the highest mutual information with $m^{(j)}$. This defines a partition over messages. For each attribute $o^{(i)}$, the associated message sub-sequence is $p_i = \{ m^{(j)} | o^{(j^*)} = o^{(i)}, \forall o^{(i)}  \}$.





Thus, given language $\mathcal{G}$, we calculate HCE as:

\begin{equation}
\text{HCE}(\mathcal{G}) = 1 - \frac{1}{n_{att}} \sum_{i=1}^{n_{att}} \frac{\entropy(o^{(i)} \mid p_i)}{\entropy(o^{(i)})}
\end{equation}

where we subtract from $1$, so that an HCE of $1$ means compositional, and $0$ means non-compositional, in alignment with other compositionality metrics, such as topsim, bosdis, posdis. To avoid confusion, we give the resulting metric a new name `Human Compositional Entropy', abbreviated as `HCE'.

HCE has similar speed advantages to posdis and bosdis, but assumes only $c_{len} \ge n_{att}$. posdis and bosdis provide alternative relaxations of residual entropy, but they both require that $c_{len} = n_{att}$.
HCE lies in $[0, 1]$, in alignment with topsim, bosdis, and posdis.
We present empirical comparisons between resent and HCE next.

\subsection{Empirical comparison of residual entropy and HCE} \label{app:resent_comparison}

\begin{table}
\small
    \centering
    \caption{Residual entropy metrics we compare}
    \begin{tabular}{p{0.2\textwidth}p{0.7\textwidth}}
    \toprule
    Metric & Description \\
    \midrule
    resent\_ours & Our implementation of residual entropy, including exhaustive search over partitions, and with optional normalization \\
    resent\_resnick & Implementation of residual entropy in code for \citet{resnick2019}, which uses a greedy approach, but requires $V=2$. It does not normalize. We generalized this to work for arbitrary $n_{att}$; and modified it to return base-2 entropy \\
    resent\_relax $=1 - \text{HCE}$ & Our relaxed version of residual entropy, works for arbitrary $n_{att}$ and $V$, optional normalization \\
    \bottomrule
    \end{tabular}
    \label{tab:resent_metrics_descriptions}
\end{table}

We compare the metrics shown in Table \ref{tab:resent_metrics_descriptions}.

resent\_ours is as far as we know a correct implementation of the residual entropy algorithm in \citet{resnick2019}. The result can optionally be normalized. Unfortunately, exhaustively searching over all possible partitions of the messages takes combinatorial time, and becomes unworkably slow for high $n_{att}$ and high $c_{len}$. resent\_resnick is our fork of the code in \footnote{https://github.com/backpropper/cbc-emecom/blob/6d01f0cdda4a8f742232b537da4f2633613f44a9/utils.py\#L164-L204}, which we have modified to work with arbitrary $n_{att}$, and to use base-2 entropy. It uses a greedy approach, but requires $V=2$, as far as we know. It does not normalize the result. resent\_relax$= 1 - HCE$ is our relaxed version of residual entropy, but without subtracting from 1. The result can optionally be normalized.

\begin{table}
\small
\caption{Comparison of metric results for  resent\_resnick and resent\_relax, using $V=2$, $n_{att}=2$, $n_{val}=5$, $c_{len}=10$. No normalization. resent\_relax$= 1 - \text{HCE}$}
\centering
\begin{tabular}{lrrr}
\toprule
\textbf{Grammar}                  & \multicolumn{1}{l}{\textbf{resent\_ours}} & \multicolumn{1}{l}{\textbf{resent\_resnick}} & \multicolumn{1}{l}{\textbf{resent\_relax}} \\
\midrule
\textsc{comp}    & 0.0000                                    & 0.0000                                       & 0.0000                                    \\
\textsc{perm}    & 0.0000                                    & 0.0000                                       & 0.0000                                    \\
\textsc{proj}    & 0.0000                                    & 0.8468                                       & 0.8333                                    \\
\textsc{shufdet} & 0.0000                                    & 0.4796                                       & 0.6452                                    \\
\textsc{shuf}    & 0.0000                                    & 0.5600                                       & 0.5389                                    \\
\textsc{rot}     & 0.0000                                    & 0.5510                                       & 0.5510                                    \\
\textsc{hol}     & 0.0000                                    & 0.5616                                       & 0.5728        \\                           
\bottomrule    
\end{tabular}
\label{tab:resent_res1}
\end{table}

\begin{table}
\small
\caption{Comparison of metric results for resent\_ours and resent\_relax, using $V=4$, $n_{att}=3$, $n_{val}=4$, $c_{len}=6$. Normalization enabled.}
\centering
\begin{tabular}{lrr}
\toprule
\textbf{Grammar}         & \multicolumn{1}{l}{\textbf{resent\_ours}} & \multicolumn{1}{l}{\textbf{resent\_relax}} \\
\midrule
\textsc{comp}   & 0.0000                     & 0.0000                       \\
\textsc{perm}         & 0.0000                     & 0.0000                       \\
\textsc{proj}      & 0.4772                     & 0.5343                       \\
\textsc{shufdet} & 0.2337                     & 0.4025                       \\
\textsc{shuf}    & 0.4340                     & 0.4973                       \\
\textsc{rot}          & 0.0814                     & 0.3867                       \\
\textsc{hol}        & 0.4954                     & 0.6183 \\
\bottomrule                    
\end{tabular}
\label{tab:resent_res2}
\end{table}

We first compare all three metrics. This requires using $V=2$, to satisfy resent\_resnick, low $n_{att}$, to keep the calculation time for resent\_ours reasonable, and high $c_{len}$ to make it possible to construct a \textsc{comp} grammar with a small $V$. We disable normalization, since resent\_resnick does not implement it. Table \ref{tab:resent_res1} shows the results, which are each averaged over 5 seeds. We can see that resent\_ours consistently scores 0, over all languages. This is probably because the utterance length is so long that there are many possible partitions, of which at least one gives zero entropy. resent\_resnick and resent\_relax give similar results, except for \textsc{shufdet} where resent\_resnick gives a lower score than resent\_relax.

Then, we increase the vocabulary size $V$. This precludes measuring resent\_resnick, which requires $V=2$, but allows for a shorter $c_{len}$ and higher $n_{att}$. We enable normalization, since both metrics support it. Table \ref{tab:resent_res2} depicts the results. In these conditions, resent\_ours is non-zero wherever resent\_relax is non-zero. resent\_relax returns results which are higher than resent\_ours, but do correlate somewhat. The values of resent\_relax for each grammar appear plausible, e.g. that for \textsc{hol} is higher than for other grammars.

\section{Example utterances} \label{app:example_utterances}

\begin{table}[htb!]
    \small
     \caption{Example utterances for 4 objects, using $n_{att}=5$ and $c_{len}=4\cdot n_{att}$.}
   \centering
    \begin{tabular}{lllllll}
\toprule 
 & \multicolumn{4}{l}{Objects} \\ 
 & (0, 0, 0, 0, 0) & (0, 0, 0, 0, 1) & (0, 0, 0, 1, 0) & (0, 0, 1, 0, 0) \\ 
\midrule 
concat & cdbacadcbcacbddacadb & cdbacadcbcacbddacadb & cdbacadcbcacbddacadb & cdbacadcbcacbddacadb \\ 
perm & dabdccbdcacabddacabc & dabdccbdcacabddacabc & dabdccbdcacabddacabc & dabdccbdcacabddacabc \\ 
rot & cbccaadbcaacdcbbddcd & cbccaadbcaacdcbbddcd & cbccaadbcaacdcbbddcd & cbccaadbcaacdcbbddcd \\ 
proj & bbdaacdaadbbabbcbbcb & bbdaacdaadbbabbcbbcb & bbdaacdaadbbabbcbbcb & bbdaacdaadbbabbcbbcb \\ 
shufdet & bcaccadcbddacdbacadb & bcaccadcbddacdbacadb & bcaccadcbddacdbacadb & bcaccadcbddacdbacadb \\ 
hol & bdabaabbacacdbaabdcd & bdabaabbacacdbaabdcd & bdabaabbacacdbaabdcd & bdabaabbacacdbaabdcd \\ 
\bottomrule 
    \end{tabular}
    \label{tab:grammar_examples_5x10}
\end{table}

Table \ref{tab:grammar_examples_5x10} depicts example utterances for $n_{att}=5$ and $c_{len} = 4 \cdot n_{att}$.

\section{CI95 values for key tables}\label{app:ci95}

Table \ref{tab:sender_5x10_ci95} shows the sender $10^5$ results, including CI95 ranges, i.e. for Table \ref{tab:sender_5x10}. Note that for any training runs that were truncated at a ratio of 20, the variance will appear to be 0, as long as all runs were truncated at a ratio of 20.

Table \ref{tab:params_change_ci95} shows the CI95 ranges for the additional results shown in Table \ref{tab:num_parameters} (the additional rows in Table \ref{tab:num_parameters} were copied from Table \ref{tab:sender_5x10}).

Table \ref{tab:zerornn_ci95} shows the CI95 ranges for the results shown in Table \ref{tab:rnnz_perm}.

Table \ref{tab:shufdet_ci95} shows the full results for the search for low \textsc{shufdet} bias, including CI95 ranges.

\begin{landscape}

\begin{table*}[htb!]
\small
\centering
\begin{tabular}{ lllllll }
\toprule
arch & params & Permute ratio & RandomProj ratio & Cumrot ratio & ShuffleWordsDet ratio & Holistic ratio \\ 
\midrule 
FC1L & 5100 & 1.12+/-0.09 & 1.9+/-0.2 & 20.000+/-0.000 & 11+/-6 & 20.000+/-0.000 \\ 
FC2L & 19428 & 1.000+/-0.000 & 2.1+/-0.2 & 20.000+/-0.000 & 7+/-3 & 20.000+/-0.000 \\ 
RNNAutoReg:LSTM & 139909 & 1.00+/-0.10 & 2.2+/-0.2 & 7+/-2 & 1.60+/-0.08 & 20.08+/-0.04 \\ 
RNNAutoReg2L:LSTM & 272005 & 0.9+/-0.2 & 1.9+/-0.3 & 5+/-1 & 1.4+/-0.2 & 20.06+/-0.04 \\ 
TransDecSoft & 272389 & 1.0+/-0.2 & 2.5+/-0.6 & 10+/-3 & 2.1+/-0.4 & 20.4+/-0.1 \\ 
TransDecSoft2L & 536965 & 1.08+/-0.07 & 2.2+/-0.5 & 10.4+/-1.0 & 2.0+/-0.1 & 20.30+/-0.06 \\ 
Hashtable & 0 & 1.0+/-0.1 & 0.98+/-0.04 & 0.98+/-0.04 & 1.02+/-0.09 & 1.1+/-0.1 \\ 
\bottomrule
\end{tabular}
\caption{CI95 ranges for Sender model results.}
\label{tab:sender_5x10_ci95}
\end{table*}

\begin{table*}[htb!]
\small
\centering
\begin{tabular}{ lllllll }
\toprule
arch & params & Permute ratio & RandomProj ratio & Cumrot ratio & ShuffleWordsDet ratio & Holistic ratio \\ 
\midrule 
RNNAutoReg:LSTM & 13195525 & 0.9+/-0.2 & 1.6+/-0.3 & 2.7+/-0.6 & 1.4+/-0.2 & 20.000+/-0.000 \\ 
FC2L & 193380 & 1.02+/-0.07 & 1.9+/-0.1 & 20.000+/-0.000 & 10+/-5 & 20.000+/-0.000 \\ 
\bottomrule
\end{tabular}
\caption{CI95 ranges for effect of number of parameters.}
\label{tab:params_change_ci95}
\end{table*}

\begin{table*}[htb!]
\small
\centering
\begin{tabular}{ lllllll }
\toprule
arch & params & Permute ratio & RandomProj ratio & Cumrot ratio & ShuffleWordsDet ratio & Holistic ratio \\ 
\midrule 
RNNAutoReg:RNN & 40837 & 0.76+/-0.09 & 2.80+/-0.10 & 18+/-3 & 1.7+/-0.1 & 20.06+/-0.04 \\ 
RNNZero:RNN & 40069 & 0.80+/-0.10 & 2.9+/-0.7 & 19+/-2 & 2.0+/-0.2 & 20.000+/-0.000 \\ 
RNNAutoReg:GRU & 106885 & 1.0+/-0.1 & 2.4+/-0.3 & 6+/-1 & 1.9+/-0.2 & 20.12+/-0.04 \\ 
RNNZero:GRU & 106117 & 1.2+/-0.2 & 2.3+/-0.3 & 4.7+/-0.5 & 1.9+/-0.1 & 20.04+/-0.04 \\ 
RNNAutoReg:LSTM & 139909 & 1.00+/-0.10 & 2.2+/-0.2 & 7+/-2 & 1.60+/-0.08 & 20.08+/-0.04 \\ 
RNNZero:LSTM & 139141 & 1.18+/-0.07 & 2.3+/-0.3 & 7+/-2 & 1.8+/-0.1 & 20.06+/-0.04 \\ 
\bottomrule
\end{tabular}
\caption{CI95 ranges for performance of ZeroRNN.}
\label{tab:zerornn_ci95}
\end{table*}

\begin{table*}[htb!]
\small
\centering
\begin{tabular}{ lllllll }
\toprule
arch & params & Permute ratio & RandomProj ratio & Cumrot ratio & ShuffleWordsDet ratio & Holistic ratio \\ 
\midrule 
RNNAutoReg:LSTM & 139909 & 1.1+/-0.1 & 2.4+/-0.3 & 6+/-1 & 1.8+/-0.1 & 20.04+/-0.03 \\ 
RNNZero:LSTM & 139141 & 1.1+/-0.1 & 2.3+/-0.2 & 5.4+/-1.0 & 1.8+/-0.1 & 20.04+/-0.03 \\ 
HierAutoReg:RNN & 64262 & 0.92+/-0.05 & 2.4+/-0.3 & 10+/-3 & 1.9+/-0.2 & 20.10+/-0.06 \\ 
HierZero:RNN & 40710 & 0.9+/-0.1 & 2.5+/-0.4 & 15+/-2 & 1.7+/-0.2 & 20.09+/-0.04 \\ 
RNNAutoReg:dgsend & 106117 & 0.92+/-0.10 & 2.3+/-0.3 & 5.5+/-0.8 & 1.5+/-0.1 & 20.13+/-0.04 \\ 
RNNZero:dgsend & 105349 & 0.92+/-0.08 & 2.1+/-0.2 & 4.6+/-0.4 & 1.5+/-0.1 & 20.11+/-0.04 \\ 
HierAutoReg:dgsend & 178566 & 1.09+/-0.07 & 1.9+/-0.2 & 6+/-1 & 1.4+/-0.1 & 20.11+/-0.04 \\ 
HierZero:dgsend & 138758 & 1.03+/-0.08 & 1.7+/-0.1 & 4.5+/-0.8 & 1.3+/-0.1 & 20.11+/-0.04 \\ 
\bottomrule
\end{tabular}
\caption{Full table for \textsc{shufdet}, including CI95 ranges.}
\label{tab:shufdet_ci95}
\end{table*}

\end{landscape}

\section{Additional results} \label{app:additional_results}

\begin{table*}[htb!]
\small
\caption{Relative compositional inductive biases for sender models, for $n_{att} = 5$, $n_{val} = 10$. $\uparrow$ and $\downarrow$ denotes columns we want to maximize or minimize respectively.}
\centering
    \begin{tabular}{ lllllllll }
    \toprule
Repr & Model & \textsc{concat} & \textsc{proj} $\downarrow$ & \textsc{pairsum} $\downarrow$ & \textsc{perm} $\downarrow$ & \textsc{rot} $\downarrow$ & \textsc{shufdet} $\uparrow$ & \textsc{hol} $\downarrow$ \\ 
\midrule 
\textsc{soft} & FC1L & 1.000 & 0.88 & 0.538 & 1.000 & \textbf{0.49} & 0.77 & 0.255 \\ 
  & FC2L & 1.000 & 0.924 & 0.54 & 1.000 & \textbf{0.49} & 0.77 & 0.253 \\ 
  & HierZero:RNN & 0.995 & 0.846 & 0.629 & 0.995 & 0.68 & 0.95 & \textbf{0.242} \\ 
  & HierAutoReg:RNN & 0.995 & 0.87 & 0.68 & 0.987 & 0.74 & \textbf{0.985} & 0.254 \\ 
  & RNNZero2L:SRU & 0.993 & \textbf{0.790} & 0.76 & \textbf{0.971} & 0.72 & 0.931 & 0.249 \\ 
  & TransDecSoft & 0.994 & 0.812 & \textbf{0.53} & 0.984 & 0.58 & 0.82 & 0.249 \\ 
  & TransDecSoft2L & 0.995 & 0.818 & 0.61 & 0.990 & 0.548 & 0.85 & 0.246 \\ 

\midrule 
\textsc{gumb} & FC1L & 0.997 & 0.846 & 0.53 & 1.000 & \textbf{0.49} & 0.750 & 0.249 \\ 
  & FC2L & 0.998 & 0.84 & \textbf{0.49} & 1.000 & \textbf{0.49} & 0.74 & 0.256 \\ 
  & HierZero:RNN & 0.991 & 0.85 & 0.64 & 0.989 & 0.72 & \textbf{0.96} & 0.249 \\ 
  & HierAutoReg:RNN & 0.994 & 0.858 & 0.63 & 0.993 & 0.68 & 0.94 & 0.256 \\ 
  & RNNZero2L:SRU & 0.992 & \textbf{0.76} & 0.722 & \textbf{0.95} & 0.691 & 0.89 & 0.255 \\ 
  & TransDecSoft & 0.993 & 0.78 & 0.53 & 0.981 & 0.53 & 0.81 & 0.251 \\ 
  & TransDecSoft2L & 0.992 & 0.784 & 0.52 & 0.979 & 0.50 & 0.77 & \textbf{0.243} \\ 

\midrule 
\textsc{discr} & FC1L & 0.961 & 0.739 & 0.423 & 0.984 & 0.46 & 0.67 & 0.252 \\ 
  & FC2L & 0.969 & 0.75 & 0.42 & 0.981 & 0.459 & 0.68 & 0.246 \\ 
  & HierZero:RNN & 0.959 & 0.731 & 0.52 & 0.93 & 0.61 & 0.70 & 0.249 \\ 
  & HierAutoReg:RNN & 0.956 & \textbf{0.65} & 0.53 & 0.94 & 0.607 & \textbf{0.74} & 0.248 \\ 
  & RNNZero2L:SRU & 0.955 & 0.67 & 0.629 & \textbf{0.87} & 0.65 & 0.715 & \textbf{0.241} \\ 
  & TransDecSoft & 0.956 & 0.68 & 0.364 & 0.93 & 0.49 & 0.63 & 0.255 \\ 
  & TransDecSoft2L & 0.951 & 0.67 & \textbf{0.304} & 0.91 & \textbf{0.42} & 0.45 & 0.245 \\ 
  
\bottomrule
\end{tabular}
\label{tab:sender_5x10_all_repr}
\end{table*}

\begin{table*}[htb!]
\small
\caption{Relative compositional inductive biases for receiver models, for $n_{att} = 5$, $n_{val} = 10$. $\uparrow$ and $\downarrow$ denotes columns we want to maximize or minimize respectively.}
\centering
    \begin{tabular}{ llllllllll }
    \toprule
Repr & Model & \textsc{concat} & \textsc{proj} $\downarrow$ & \textsc{pairsum} $\downarrow$ & \textsc{perm} $\downarrow$ & \textsc{rot} $\downarrow$ & \textsc{shuf} $\uparrow$ & \textsc{shufdet} $\uparrow$ & \textsc{hol} $\downarrow$ \\ 

\midrule 
\textsc{soft} & CNN & 0.997 & 0.57 & 0.59 & 0.965 & 0.92 & 0.63 & 0.82 & 0.106 \\ 
  & FC2L & 1.000 & 0.77 & 0.50 & 1.000 & \textbf{0.44} & 0.63 & 0.84 & 0.101 \\ 
  & Hier:GRU & 0.996 & 0.57 & 0.51 & 0.95 & 0.94 & \textbf{0.972} & 0.970 & 0.111 \\ 
  & Hier:dgrecv & 0.993 & \textbf{0.52} & 0.43 & 0.972 & 0.96 & 0.89 & 0.91 & 0.097 \\ 
  & RNN1L:LSTM & 0.994 & 0.529 & 0.42 & 0.93 & 0.83 & 0.721 & 0.88 & \textbf{0.091} \\ 
  & RNN1L:dgrecv & 0.993 & 0.532 & \textbf{0.38} & 0.963 & 0.90 & 0.70 & 0.91 & 0.096 \\ 
  & RNN2L:GRU & 0.996 & 0.568 & 0.57 & 0.93 & 0.94 & 0.90 & \textbf{0.973} & 0.112 \\ 
  & RNN2L:SRU & 0.994 & 0.55 & 0.47 & \textbf{0.91} & 0.89 & 0.90 & 0.962 & 0.099 \\ 
\midrule 
\textsc{gumb} & CNN & 0.994 & 0.53 & 0.49 & 0.93 & 0.79 & 0.58 & 0.85 & 0.107 \\ 
  & FC2L & 1.000 & 0.657 & 0.37 & 1.000 & \textbf{0.38} & 0.485 & 0.75 & 0.107 \\ 
  & Hier:GRU & 0.994 & 0.54 & 0.46 & 0.96 & 0.943 & \textbf{1.000} & 0.999 & 0.099 \\ 
  & Hier:dgrecv & 0.998 & 0.54 & 0.50 & 0.963 & 0.982 & \textbf{1.000} & \textbf{1.000} & 0.096 \\ 
  & RNN1L:LSTM & 0.997 & 0.511 & 0.30 & 0.977 & 0.84 & 0.85 & 0.937 & \textbf{0.094} \\ 
  & RNN1L:dgrecv & 0.996 & 0.62 & 0.61 & 1.000 & 0.95 & 0.88 & 0.93 & 0.103 \\ 
  & RNN2L:GRU & 0.995 & 0.54 & 0.42 & 0.95 & 0.95 & 0.94 & 0.981 & 0.10 \\ 
  & RNN2L:SRU & 0.995 & \textbf{0.45} & \textbf{0.28} & \textbf{0.87} & 0.80 & 0.947 & 0.967 & 0.100 \\ 
\midrule 
\textsc{discr} & CNN & 0.8 & \textbf{0.28} & 0.32 & \textbf{0.7} & 0.59 & 0.5 & 0.70 & 0.098 \\ 
  & FC2L & 0.93 & 0.67 & 0.44 & 0.96 & \textbf{0.47} & 0.48 & 0.69 & 0.103 \\ 
  & Hier:GRU & 0.973 & 0.49 & 0.43 & 0.981 & 0.8 & \textbf{1.000} & \textbf{0.98} & 0.097 \\ 
  & Hier:dgrecv & 0.96 & 0.508 & 0.47 & 0.976 & 0.6 & 0.7 & 0.93 & 0.096 \\ 
  & RNN1L:LSTM & 0.989 & 0.44 & 0.25 & 0.90 & 0.568 & 0.77 & 0.90 & 0.100 \\ 
  & RNN1L:dgrecv & 0.93 & 0.59 & 0.5 & 0.991 & 0.87 & 0.989 & 0.95 & 0.120 \\ 
  & RNN2L:GRU & 0.986 & 0.46 & 0.37 & 0.93 & 0.80 & 0.92 & \textbf{0.98} & 0.100 \\ 
  & RNN2L:SRU & 0.961 & 0.36 & \textbf{0.21} & 0.77 & 0.60 & 0.82 & 0.85 & \textbf{0.094} \\ 
 
\bottomrule
\end{tabular}
\label{tab:receiver_5x10_all_repr}
\end{table*}

Tables \ref{tab:sender_5x10_all_repr} and \ref{tab:receiver_5x10_all_repr} show more results for both sender and receiver models, trained supervised in isolation. We are using the evaluation method here of measuring the number of steps to train \textsc{concat} to $\acc_{tgt} = 0.95$, then train other grammars for the same number of steps, then report $\acc_{train}$ for each of these other grammars. Each result is an averaged over 3 runs.

In addition, we experimented in this table with using other loss functions than cross-entropy: we experiment with adding a Gumbel sampler to the network output, prior to the loss function (\textsc{gumb}); and adding a stochastic sampler to the network output, and train using REINFORCE (\textsc{discr}) (i.e. `discrete').

\begin{table*}[htb!]
\small
\caption{Example results for measuring compositional inductive bias for end to end architectures, for $n_{att} = 5$, $n_{val} = 10$. $\uparrow$ and $\downarrow$ denotes columns we want to maximize or minimize respectively.}
\centering
    \begin{tabular}{ llllllll }

    \toprule
Repr & Send arch & Recv arch & \textsc{concat} $\uparrow$ & \textsc{perm} $\downarrow$ & \textsc{rot} $\downarrow$ & \textsc{shufdet} $\uparrow$ & \textsc{pairsum} $\downarrow$ \\ 
\midrule 
\textsc{soft} & RNNAutoReg:dgsend & RNN:dgrecv & 0.919 & 0.859 & \textbf{0.33} & 0.893 & \textbf{0.61} \\ 
  & RNNAutoReg:GRU & RNN:GRU & \textbf{0.949} & 0.94 & 0.89 & \textbf{0.976} & 0.73 \\ 
  & HierZero:GRU & Hier:GRU & 0.85 & \textbf{0.80} & 0.53 & 0.74 & 0.63 \\ 
  & RNNZero2L:RNN & RNN2L:RNN & 0.870 & \textbf{0.80} & 0.83 & 0.954 & 0.66 \\ 
  & RNNZero:RNN & RNN:RNN & 0.87 & 0.829 &  & 0.946 & 0.72 \\ 
  & RNNZero:GRU & RNN:GRU & 0.887 & 0.88 & 0.891 & 0.913 & 0.78 \\ 
\midrule 
\textsc{gumb} & RNNAutoReg:dgsend & RNN:dgrecv & 0.46 & 0.38 & 0.32 & 0.25 & \textbf{0.096} \\ 
  & RNNAutoReg:GRU & RNN:GRU & 0.381 & 0.433 & 0.388 & 0.344 & 0.350 \\ 
  & HierZero:GRU & Hier:GRU & 0.36 & 0.7 & \textbf{0.26} & 0.202 & 0.27 \\ 
  & RNNZero2L:RNN & RNN2L:RNN & \textbf{0.935} & 0.28 & 0.48 & 0.82 & 0.492 \\ 
  & RNNZero:RNN & RNN:RNN & 0.80 & \textbf{0.25} & 0.501 & 0.56 &  \\ 
  & RNNZero:GRU & RNN:GRU & 0.909 & 0.85 & 0.746 & \textbf{0.89} & 0.51 \\ 
\midrule 
\textsc{discr} & RNNAutoReg:dgsend & RNN:dgrecv & 0.18 & \textbf{0.30} & \textbf{0.262} & 0.27 & \textbf{0.17} \\ 
  & RNNAutoReg:GRU & RNN:GRU & 0.36 & 0.68 & 0.34 & 0.38 & 0.50 \\ 
  & HierZero:GRU & Hier:GRU & 0.40 & 0.34 & 0.31 & 0.34 & 0.36 \\ 
  & RNNZero2L:RNN & RNN2L:RNN & 0.72 & 0.56 & 0.42 & 0.28 & 0.38 \\ 
  & RNNZero:RNN & RNN:RNN & \textbf{0.827} & 0.750 &  & 0.279 & 0.282 \\ 
  & RNNZero:GRU & RNN:GRU & 0.58 & 0.80 & 0.445 & \textbf{0.61} & 0.577 \\  
\bottomrule
\end{tabular}
\label{tab:e2e_5x10_all_repr}
\end{table*}

Table \ref{tab:e2e_5x10_all_repr} shows additional results for end to end training from models first pre-trained supervised on specific grammars. The methodology used to generate these tables was:

\begin{itemize}
    \item train the sender and a receiver model supervised until they achieve $\acc_{tgt}$ on $\mathcal{G}$
    \item place the sender and receiver end to end, as an auto-encoder
    \item train the auto-encoder end-to-end for $T$ steps
    \item measure the accuracy of either the sender or the receiver model on the original language $\mathcal{G}$
\end{itemize}

The results are depicted in Table \ref{tab:e2e_5x10_all_repr}, using $T = 10,000$. Results are each averaged over three runs.

\section{\textsc{shuf} and \textsc{shufdet} grammar and additional results} \label{app:shuf}.


We would like to encourage the models to emerge relocatable atomic groups of tokens, that is something similar to words in natural language. We thus create two artificial grammars, which we would like neural models to acquire quickly: \textsc{shuf} and \textsc{shufdet}. \textsc{shuf} (`shuffle') permutes the order sub-messages $w$, prior to concatenation. The permutation is sampled uniformly once per utterance in the language. For example, if object (red,box) maps to utterance‘adaaccad’, then after permutation of the word order, the shuffled message could be ‘ccadadaa‘, equivalent to ‘boxred’ in English. \textsc{shufdet} (`shuffle deterministically') samples one permutation for each value of the first attribute of the utterance. Thus the permutation is deterministic, given the value of the first attribute, and the sampled permutations.

In natural language, whilst it is not the case that all sentences can be permuted without changing the meaning, it is the case that many sentences can be re-arranged, without much affecting a human's understanding.

For a Sender, evaluating on \textsc{shuf} is not reasonable, since there is no obvious way for the Sender to know which order we are evaluating on. Hence, \textsc{shufdet} might be reasonable for a Sender model. In natural language, some word orders are dependent on the values of certain words. For example, in French, the adjective `neuve' follows a noun, whereas `nouvelle' precedes it.

\textsc{shuf} and \textsc{shufdet} contrast with the other artificial grammars we propose in that we feel that models with a similar compositional inductive bias to humans should acquire these grammars quickly.

\begin{table}
\small
\caption{Comparison of various RNN models, searching for low \textsc{shufdet}. Results are mean over 10 runs.}
\begin{tabular}{lrrrrrl}
\textbf{Model}     & \multicolumn{1}{l}{\textbf{Params}} & \multicolumn{1}{l}{\textbf{\textsc{perm}}} & \multicolumn{1}{l}{\textbf{\textsc{proj}}} & \multicolumn{1}{l}{\textbf{\textsc{rot}}} & \multicolumn{1}{l}{\textbf{\textsc{shufdet}}} & \textbf{\textsc{hol}}        \\
LSTM    & 139909                     & \cellcolor[HTML]{D9EAD3}1.08          & \cellcolor[HTML]{D9EAD3}2.4          & \cellcolor[HTML]{D9EAD3}6.3      & \cellcolor[HTML]{D9EAD3}1.8           & \cellcolor[HTML]{F4CCCC}> 20 \\
LSTM-Z       & 139141                     & \cellcolor[HTML]{D9EAD3}\textbf{1.15} & \cellcolor[HTML]{D9EAD3}2.3          & \cellcolor[HTML]{D9EAD3}5.4      & \cellcolor[HTML]{D9EAD3}1.83          & \cellcolor[HTML]{F4CCCC}> 20 \\
dgsend  & 106117                     & \cellcolor[HTML]{D9EAD3}0.92          & \cellcolor[HTML]{D9EAD3}2.3          & \cellcolor[HTML]{D9EAD3}5.5      & \cellcolor[HTML]{D9EAD3}1.52          & \cellcolor[HTML]{F4CCCC}> 20 \\
dgsend-Z     & 105349                     & \cellcolor[HTML]{D9EAD3}0.92          & \cellcolor[HTML]{D9EAD3}2.1          & \cellcolor[HTML]{D9EAD3}4.6      & \cellcolor[HTML]{D9EAD3}1.54          & \cellcolor[HTML]{F4CCCC}> 20 \\
HUSendA:RNN    & 64262                      & \cellcolor[HTML]{D9EAD3}0.92          & \cellcolor[HTML]{D9EAD3}2.4          & 10                               & \cellcolor[HTML]{D9EAD3}1.9           & \cellcolor[HTML]{F4CCCC}> 20 \\
HUSendZ:RNN       & 40710                      & \cellcolor[HTML]{D9EAD3}0.93          & \cellcolor[HTML]{D9EAD3}\textbf{2.5} & \textbf{15}                      & \cellcolor[HTML]{D9EAD3}1.68          & \cellcolor[HTML]{F4CCCC}> 20 \\
HUSendA:dgsend & 178566                     & \cellcolor[HTML]{D9EAD3}1.09          & \cellcolor[HTML]{D9EAD3}1.9          & \cellcolor[HTML]{D9EAD3}5.8      & \cellcolor[HTML]{D9EAD3}1.4           & \cellcolor[HTML]{F4CCCC}> 20 \\
HUSendZ:dgsend    & 138758                     & \cellcolor[HTML]{D9EAD3}1.03          & \cellcolor[HTML]{D9EAD3}1.67         & \cellcolor[HTML]{D9EAD3}4.5      & \cellcolor[HTML]{D9EAD3}\textbf{1.27} & \cellcolor[HTML]{F4CCCC}> 20
\end{tabular}
\label{tab:shufdet_results_full}
\end{table}

Table \ref{tab:shufdet_results_full} shows more complete results for our comparison of \textsc{shufdet} bias across various recurrent neural architectures.

\section{End-to-end training}\label{app:endtoend}



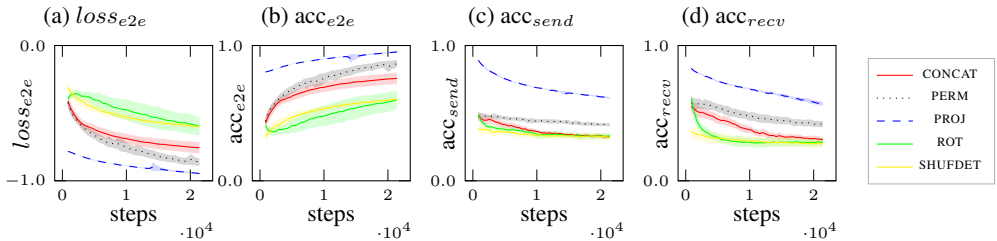
\begin{figure*}[htb!]
\centering
\foreach \valuecolumn/\valuedesclong\miny/\maxy/\ytickdist in {e2e_loss/$loss_{e2e}$/-1.0/0.0/1.0, e2e_acc/$\acc_{e2e}$/0/1.0/1.0, send_acc/$\acc_{send}$/0/1.0/1.0, recv_acc/$\acc_{recv}$/0/1.0/1.0} {
\begin{subfigure}{0.19\textwidth}
\caption{\small{\valuedesclong}}
\vspace{-0.2cm}
\begin{tikzpicture}
    \begin{axis}[
        small,
        scale only axis,
        width=2.1cm,
        height=1.8cm,
        ylabel=\valuedesclong,
        xlabel=steps,
        ylabel shift=-0.6cm,
        xlabel shift=-0.15cm,
        yticklabel style={
            font=\tiny,
            /pgf/number format/fixed,
            /pgf/number format/precision=1,
            /pgf/number format/fixed zerofill
        },
        xticklabel style={
            font=\tiny
        },
        scaled y ticks=false,
        ytick distance=\ytickdist,
        mark size=1pt,
        ymin=\miny,
        ymax=\maxy,
    ]
    \foreach \grammar/\color/\linetype in {comp/red/solid, perm/black/dotted, proj/blue/dashed, rot/green/solid, shufdet/yellow/solid} {
        \edef\tmp{\noexpand\addplot [
            name path=lower,
            draw=none,
        ] table [
            x=step,
            y expr=\noexpand\thisrow{\valuecolumn} - \noexpand\thisrow{\valuecolumn_ci95},
            discard if not={grammar}{\grammar},
            col sep=comma,
        ] {data/training_curves.csv}};
        \tmp;
        \edef\tmp{\noexpand\addplot [
            name path=upper,
            draw=none,
        ] table [
            x=step,
            y expr=\noexpand\thisrow{\valuecolumn} + \noexpand\thisrow{\valuecolumn_ci95},
            discard if not={grammar}{\grammar},
            col sep=comma,
        ] {data/training_curves.csv}};
        \tmp;
        \edef\tmp{\noexpand\addplot[\color!80, fill opacity=0.2] fill between[of=lower and upper];}
        \tmp;
        \edef\tmp{\noexpand\addplot [
            \color,
            \linetype,
        ] table [
            x=step,
            y expr=\noexpand\thisrow{\valuecolumn},
            discard if not={grammar}{\grammar},
            col sep=comma,
        ] {data/training_curves.csv}};
        \tmp;
    }
    \end{axis}
\end{tikzpicture}
\end{subfigure}
}
\begin{subfigure}{0.16\textwidth}
\centering
\begin{tikzpicture}
    \begin{customlegend}[
        tiny,
        legend entries={\textsc{concat},\textsc{perm},\textsc{proj},\textsc{rot},\textsc{shufdet}},
        legend style={
            draw=black!30,
            /tikz/every even column/.append style={column sep=0.3cm},
        },
    ]
    \addlegendimage{red}
    \addlegendimage{black,dotted}
    \addlegendimage{blue,dashed}
    \addlegendimage{green,solid}
    \addlegendimage{yellow,solid}
    \end{customlegend}
\end{tikzpicture}
\end{subfigure}
\vspace{-0.2cm}
\caption{Training curves for LSTM sender and LSTM receiver placed end-to-end after supervised training on the specified grammar. Each curve is mean over 10 runs, and shading is CI95. $n_{att}=5$.}
\label{fig:e2e_training_curves}
\end{figure*}


We experimented with training a Sender and Receiver model supervised on a specific grammar, placing end-to-end, and continuing training, using REINFORCE. Figure \ref{fig:e2e_training_curves} shows the results for an LSTM Sender and Receiver. We see clear differences between the grammars, but some are surprising. We expected that \textsc{concat} and \textsc{perm} would have the smallest $loss_{e2e}$ and the best $\acc_{e2e}$, but \textsc{proj} did better, and \textsc{perm} did better than \textsc{concat}. $\acc_{send}$ and $\acc_{recv}$ measures the accuracy of the emergent language w.r.t. the original grammar. We thought that \textsc{concat} and \textsc{perm} would deviate least, but \textsc{proj} deviated the least, for reasons unclear. We feel that this scenario might provide opportunities to investigate generalization and exploration under controlled conditions.


\section{Acquisition accuracy given fixed training budget}\label{app:constant_steps}

\begin{table}
\small
\caption{Compositional inductive bias estimated by accuracy after training for fixed number of steps. In red, results below 0.5, and in green, results over 0.8. $\acc_{tgt}$ for \textsc{concat} is 0.99. Results are averaged over 5 seeds.}
\begin{tabular}{p{0.4\textwidth}rrrrrr}
\textbf{Model}                                           & \multicolumn{1}{l}{\textbf{Params}} & \multicolumn{1}{l}{\textbf{\textsc{perm}}} & \multicolumn{1}{l}{\textbf{\textsc{proj}}} & \multicolumn{1}{l}{\textbf{\textsc{rot}}} & \multicolumn{1}{l}{\textbf{\textsc{shufdet}}} & \multicolumn{1}{l}{\textbf{\textsc{hol}}} \\
1-layer MLP                & 5100                                & \cellcolor[HTML]{D9EAD3}0.995              & \cellcolor[HTML]{D9EAD3}0.848              & \cellcolor[HTML]{F4CCCC}0.478             & 0.741                                         & \cellcolor[HTML]{F4CCCC}0.258             \\
2-layer MLP                                              & 19428                               & \cellcolor[HTML]{D9EAD3}0.995              & 0.78                                       & \cellcolor[HTML]{F4CCCC}0.481             & 0.715                                         & \cellcolor[HTML]{F4CCCC}0.252             \\
1-layer LSTM                  & 139909                              & \cellcolor[HTML]{D9EAD3}0.967              & \cellcolor[HTML]{D9EAD3}0.823              & 0.721                                     & \cellcolor[HTML]{D9EAD3}0.894                 & \cellcolor[HTML]{F4CCCC}0.251             \\
2-layer LSTM                                             & 272005                              & \cellcolor[HTML]{D9EAD3}0.961              & \cellcolor[HTML]{D9EAD3}0.817              & 0.705                                     & \cellcolor[HTML]{D9EAD3}0.9                   & \cellcolor[HTML]{F4CCCC}0.249             \\
\end{tabular}
\label{tab:constant_steps}
\end{table}

Table \ref{tab:sender_5x10} is conveniently intuitive to read, however the number of steps to reach convergence is unbounded, and some combinations of model and grammar might never converge. We worked around this issue by stopping training at $b = 20$. An alternative approach is to train for a fixed number of training steps, and report the resulting accuracy. For each model, we train \textsc{concat} until $\acc_{tgt}$, and then train other grammars for the same number of steps. Table \ref{tab:constant_steps} shows results for some of the architectures from Table \ref{tab:sender_5x10}. An obvious downside is that we cannot tell which grammars will ever converge. However, the measures are squashed conveniently into $[0, 1]$, and the experiments are much faster to run (constant time given convergence time of \textsc{concat}).
We can see that the relative accuracies correlate well with the results in Table \ref{tab:sender_5x10}.

\section{Human evaluation}\label{app:turk}

\begin{figure}
    \centering
    \includegraphics[width=0.45\columnwidth]{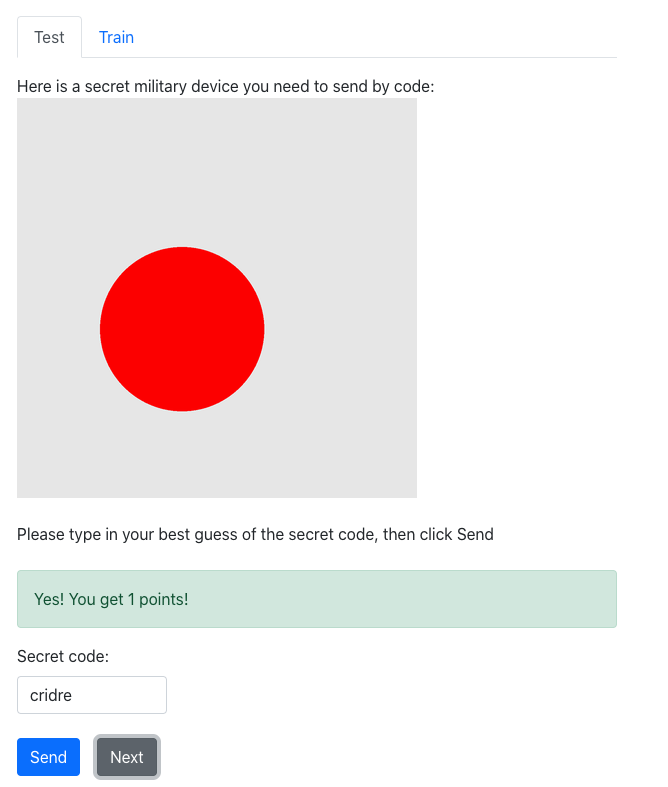}
    \caption{User interface for our `Secret Spy Codes' game. This image depicts a \textsc{perm} grammar, where the original utterance was `redcir'.}
    \label{fig:mturk_full}
\end{figure}

We constructed an MTurk \citep{mturk} task, `Secret Spy Codes', in order to evaluate human performance on \textsc{Icy} grammars. Figure \ref{fig:mturk_full} shows the user interface. Human subjects are tasked with writing out the code that represents depicted geometric objects.
A challenge we found with human experiments was that humans need substantial effort in order to learn just a few new words.
Thus, we use objects with only two attributes: shape and color.
We considered two scenarios, which we depict as `eng' and `synth'.

\subsection{Dataset}

`synth' uses artificially generated random words for each attribute value. We sample 2-letter words from a vocabulary size of 4. Each utterance therefore has 4 letters: 2 for each of shape and color. Empirically, humans found these words challenging to remember, so we used just three possible values for each attribute. Thus, there were 9 combinations in total.

`eng' uses 3-letter English abbreviations for attribute values, e.g. `tri' for `triangle', and `grn' for `green'. The words describing each attribute value in `eng' are relatively easy for a human to learn. Therefore, we used 5 attribute values for each attribute, giving 25 possible combinations.

Subjects had access to a `training' panel, where they could cycle through example images and utterances for the current grammar, then switch to a `test' panel to enter their answer. Thus, subjects could obtain a perfect score, given sufficient time to browse through the training examples. However, we hoped that the time required to browse through the training examples would vary depending on how easy the grammar was to memorize.

We held out three color-shape combinations, that were not made available in the training panel. For example, subjects might have access to a red circle and a blue triangle, but not a red triangle. Subjects who could perceive the compositional structure of a grammar should be able to get these holdout instances correct.

\subsection{Payment, incentives and curriculum}

We paid subjects to play games comprising 50 examples, where each game uses single grammar instance. The game provided immediate points rewards, and sound effects for right and wrong answers. We received feedback such as `good', `an interesting task', and `It is very interesting task, kindly upload more tasks like this in future', which suggested that subjects enjoyed playing.

Payment was a base rate plus a linear function of the subject's total score. We found that paying only a base rate worked well initially, but as we ran more tasks, subjects quickly learned to just put random utterances for each example, completing quickly, and scoring 0. Paying a linear function of the subject's total score solved this issue. We paid a base rate in order that some of the harder tasks were not too discouraging.

To avoid overwhelming subjects with learning many new utterances at the start, we start the game with only two color-shape combinations, and add one additional combination every 8 test examples. Subjects have buttons to add and remove color-shape combinations, so they can control their own curriculum. To incentivize subjects to increase the number of color-shape combinations, the score for each example is linearly proportional to the number of color-shape combinations available.

\subsection{Scoring}

Subjects were given points for each correct answer. The points for each example was calculated as (number available objects at that time) - 1. For \textsc{eng} dataset, the maximum possible score is this $(25 - 3 - 1) * 50 = 1050$ (we remove 3 objects for holdout; holdout itself scores identically to other examples), while for \textsc{synth} dataset, the maximum possible score is $(9 - 3 - 1) * 50 = 250$.

If someone uses the default curriculum, without modifying the number of available cards, then the maximum possible score is $1 * 8 + 2 * 8 + 3 * 8 + 4 * 8 + 5 * 8 + 10 * 6 = 180$, independent of dataset.

\subsection{Acceptance criteria}

We automatically paid all workers who earned at least a score of 100. We automatically rejected payment to all workers who scored 0. In between these two values, we inspected manually. Anyone who gave the same answer, or almost the same answer, for all examples, we rejected payment for, otherwise we accepted. We noticed that the typical score for anyone putting the same answer for all examples was around 41, which corresponded to the score at chance in this scenario.

For our results tables, we include everyone who scored above 50, and ignore results for anyone who scored below 50.

\subsection{Evaluation}

We measured subjects performance in two ways: across all test examples, and uniquely on the 3 held out examples.

\begin{table*}[htb!]
\small
\centering
\begin{tabular}{ lllll }
\toprule
Grammar & N & $t$ (mins) & score & $\text{acc}_{holdout}$ \\ 
\midrule 
\textsc{comp} & 18 & $17\pm4$ & $140\pm10$ & $0.2\pm0.1$ \\ 
\textsc{perm} & 17 & $22\pm7$ & $160\pm10$ & $0.04\pm0.05$ \\ 
\textsc{proj} & 15 & $16\pm3$ & $140\pm10$ & $0.02\pm0.04$ \\ 
\textsc{rot} & 18 & $21\pm6$ & $140\pm10$ & $0.04\pm0.07$ \\ 
\textsc{shufdet} & 17 & $19\pm5$ & $140\pm10$ & $0.06\pm0.06$ \\ 
\bottomrule
\end{tabular}
\caption{Human evaluation results, \textsc{synth} dataset}
\label{tab:turk_synth}
\end{table*}

\begin{table*}[htb!]
\small
\centering
\begin{tabular}{ lllll }
\toprule
Grammar & N & $t$ (mins) & score & $\text{acc}_{holdout}$ \\ 
\midrule 
\textsc{comp} & 15 & $14\pm3$ & $200\pm100$ & $0.6\pm0.2$ \\ 
\textsc{perm} & 18 & $30\pm10$ & $300\pm100$ & $0.2\pm0.2$ \\ 
\textsc{proj} & 17 & $19\pm4$ & $160\pm40$ & $0.04\pm0.08$ \\ 
\textsc{rot} & 21 & $24\pm5$ & $170\pm30$ & $0.02\pm0.03$ \\ 
\textsc{shufdet} & 17 & $20\pm8$ & $240\pm80$ & $0.7\pm0.2$ \\ 
\bottomrule
\end{tabular}
\caption{Human evaluation results, \textsc{eng} dataset}
\label{tab:turk_eng}
\end{table*}

Tables \ref{tab:turk_synth} and \ref{tab:turk_eng} show the results. Analysis of $\acc_{holdout}$ is already included in the main paper body. As far as score and timings, the subjects always have access to a `training' tab, where they can view the code for all objects except the holdout objects, therefore it is possible to obtain a perfect score in all scenarios, by referring to the training objects. We decided that it was better to provide a self-service training table, than to simply be measuring who had decided to write down a translation table between object and code, e.g. on a piece of paper. However, both our provision of a training table, and the possibility that subjects write down a translation table, means that there is a negligible difference in scores across grammars, on the training data.

\end{document}